\documentclass[11pt, a4paper, logo, copyright]{hail_lab}

\usepackage[utf8]{inputenc} 
\usepackage[T1]{fontenc}    
\usepackage{url}            
\usepackage{booktabs}       
\usepackage{amsfonts}       
\usepackage{nicefrac}       
\usepackage{microtype}      
\usepackage{xcolor}         

\usepackage{amsmath,amsfonts,bm}

\def\eqref#1{equation~\ref{#1}}

\def\1{\bm{1}}

\DeclareMathAlphabet{\mathsfit}{\encodingdefault}{\sfdefault}{m}{sl}
\SetMathAlphabet{\mathsfit}{bold}{\encodingdefault}{\sfdefault}{bx}{n}

\usepackage{microtype}
\usepackage{graphicx}
\graphicspath{{figures/}}
\usepackage{booktabs}
\usepackage{amsmath,bm}
\usepackage{amsthm}
\usepackage[numbers]{natbib}
\usepackage{enumerate}
\usepackage{wrapfig}
\usepackage{color, soul}
\usepackage{psfrag}
\usepackage{adjustbox}
\usepackage{xspace}
\usepackage{amssymb}
\usepackage{multirow}
\usepackage{afterpage}
\usepackage{colortbl}
\usepackage{float}
\usepackage{textcomp}
\usepackage{siunitx}
\usepackage{mdframed}
\usepackage{makecell}
\usepackage[super]{nth}

\usepackage{enumitem}
\usepackage{bbding}
\usepackage{pifont}
\usepackage{bbm}
\usepackage{xcolor}
\usepackage{pdflscape}
\usepackage{theoremref}
\usepackage{pifont}
\usepackage{adjustbox}
\usepackage{xcolor}
\usepackage{mathtools}
\usepackage{subcaption}
\usepackage{pbox}
\usepackage{longtable}
\usepackage{listings}
\usepackage[nomath]{lmodern}

\colorlet{darkgreen}{green!65!black}
\colorlet{darkblue}{blue!75!black}
\colorlet{darkred}{red!80!black}
\definecolor{statistical}{HTML}{8c564b}
\definecolor{structural}{HTML}{0070C0}
\definecolor{semantic}{HTML}{008080}
\definecolor{yellow}{HTML}{f7c600}
\definecolor{lightblue}{HTML}{0071bc}
\definecolor{lightgreen}{HTML}{39b54a}
\definecolor{deemph}{gray}{0.55}

\definecolor{baselinecolor}{gray}{.95}
\definecolor{graycolor}{gray}{.95}

\usepackage[pagebackref=false, breaklinks=true, colorlinks, citecolor=lightblue, urlcolor=lightblue, bookmarks=false]{hyperref}

\newlength\savewidth
\newcolumntype{x}[1]{>{\centering\arraybackslash}p{#1pt}}
\newcolumntype{y}[1]{>{\raggedright\arraybackslash}p{#1pt}}
\newcolumntype{z}[1]{>{\raggedleft\arraybackslash}p{#1pt}}

\definecolor{textgreen}{RGB}{57, 172, 57}
\definecolor{textred}{RGB}{200, 10, 10}
\definecolor{boxyellow}{HTML}{FAF5E6}
\definecolor{frameyellow}{HTML}{B7950B}
\definecolor{boxpurple}{HTML}{F4EFF6}
\definecolor{framepurple}{HTML}{6C3483}
\definecolor{boxblue}{HTML}{EEF4F8}
\definecolor{frameblue}{HTML}{2874A6}
\definecolor{boxgray}{HTML}{F0F2F3}
\definecolor{framegray}{HTML}{5D6D7E}
\definecolor{boxgreen}{HTML}{EAFaf1}
\definecolor{framegreen}{HTML}{196F3D}
\usepackage{pifont} 
\newcommand{\cmark}{\textcolor{textgreen}{\ding{51}}}
\newcommand{\xmark}{\textcolor{textred}{\ding{55}}}

\usepackage[most]{tcolorbox}
\tcbset{
    center title,
    left=0pt,
    right=0pt,
    top=0pt,
    bottom=0pt,
    colframe=black,
    colback=gray!10,
    width=\linewidth,
    enlarge left by=0mm,
    boxsep=5pt,
    boxrule=1pt,
    arc=0pt,outer arc=0pt,
}

\newtcolorbox{promptbox}[1][]{
    enhanced,
    colback=white,
    colframe=black,
    fonttitle=\bfseries,
    title=Prompt,
    attach boxed title to top left={xshift=10pt, yshift*=-\tcboxedtitleheight/2},
    boxed title style={colback=black},
    top=12pt, bottom=10pt, left=10pt, right=10pt,
    #1
}

\tcbset{
    fancybox/.style={
        enhanced,
        breakable,
        arc=1mm,
        outer arc=1mm,
        boxrule=0pt,
        leftrule=1mm,
        toptitle=0.5mm,
        bottomtitle=0.5mm,
        fonttitle=\bfseries\sffamily\small,
        fontupper=\scriptsize,
        coltitle=white,
        drop shadow
    }
}

\newtcolorbox{thoughtbox}{
    fancybox,
    colback=boxyellow,
    colframe=frameyellow,
    coltitle=black,
    title=Thought
}
\newtcolorbox{userbox}{
    fancybox,
    colback=boxpurple,
    colframe=framepurple,
    title=User
}
\newtcolorbox{agentbox}{
    fancybox,
    colback=boxblue,
    colframe=frameblue,
    title=Agent
}
\newtcolorbox{outputbox}{
    fancybox,
    colback=boxgray, 
    colframe=framegray,
    coltitle=black,
    title=Execution Output
}
\newtcolorbox{solutionbox}{
    fancybox,
    colback=boxgreen,
    colframe=framegreen,
    title=Solution
}

\definecolor{codegreen}{rgb}{0.0, 0.5, 0.0}
\definecolor{codegray}{rgb}{0.4, 0.4, 0.4}
\definecolor{codepurple}{rgb}{0.50, 0, 0.50}
\definecolor{backcolour}{rgb}{0.97, 0.97, 0.97}

\lstdefinestyle{mystyle}{
    backgroundcolor=\color{backcolour},
    commentstyle=\color{codegreen},
    keywordstyle=\color{magenta},
    stringstyle=\color{codepurple},
    basicstyle=\ttfamily\scriptsize, 
    breakatwhitespace=false,
    breaklines=true,
    captionpos=b,
    keepspaces=true,
    numbers=none,              
    showspaces=false,
    showstringspaces=false,
    showtabs=false,
    tabsize=2,
    frame=single,
    rulecolor=\color{black!10}, 
    frameround=fttt,            
    upquote=true
}
\newcommand{\ours}{\texttt{OpenSLA}\xspace}
\newcommand{\oursbase}{\texttt{OpenSLA-B}\xspace}
\newcommand{\ourshq}{\texttt{OpenSLA-H}\xspace}

\newcommand{\gpt}{{GPT-5.6-Luna}\xspace}
\newcommand{\deepseek}{{DeepSeek-V3.2}\xspace}

\newcommand{\numaction}{60\xspace}
\newcommand{\nummodality}{79\xspace}
\newcommand{\numpatient}{116K\xspace}

\definecolor{observation}{HTML}{2E3192}
\definecolor{relational}{HTML}{f7931e}
\definecolor{global}{HTML}{006837}
\definecolor{action}{HTML}{662d91}

\newcommand{\numdataset}{six\xspace}

\title{
Sensor-Language-Action Models
}

\author[1*]{Yuekai Xu}
\author[1*]{Zitao Shuai}
\author[1$\dagger$]{Yuzhe Yang}

\affil[1]{University of California, Los Angeles}
\correspondingauthor{ \textsuperscript{*}Equal contribution (co-first authors). \quad $^\dagger$Correspondence to: yuzhey@ucla.edu.}

\datasetlink{https://hf.co/yang-ai-lab/OpenSLA}
\codelink{https://github.com/yang-ai-lab/OpenSLA}
\projecturl{https://yang-ai-lab.github.io/OpenSLA}

\begin{abstract}
\vspace{-5pt}

Sensors are useful not only for understanding the world but also for deciding what to do \textit{next}.
Existing sensor models however largely stop at \textit{perception}: they recognize states or predict outcomes, leaving \textit{actions} modeled separately through task-specific and often closed label spaces.
We introduce \textbf{Sensor-Language-Action (SLA)} modeling, a framework that connects multimodal sensor observations, natural language, and actions within a unified model.
SLA uses language as a semantic interface between sensing and acting, allowing heterogeneous actions to be represented, predicted, and explained while remaining grounded in the underlying sensor evidence.
We build a large-scale SLA benchmark consisting of datasets that span more than {116,000} individuals, 79 sensor modalities, and \numaction action groups, together with a multi-faceted captioning pipeline that aligns user context, sensor dynamics, and action evidence.
Building on this framework, we present \ours, a unified SLA model for hierarchical action prediction, state understanding, and action explanation.
Extensive experiments on real-world tasks in clinical prediction, operating rooms, and metabolic health verify its superior performance over the state-of-the-art.
\ours also demonstrates intriguing capabilities including language-guided evidence grounding and zero-shot generalization to unseen actions and cohorts.

\end{abstract}

\begin{document}

\maketitle
\afterpage{\begin{figure}[H]
    \centering
    \includegraphics[width=\linewidth]{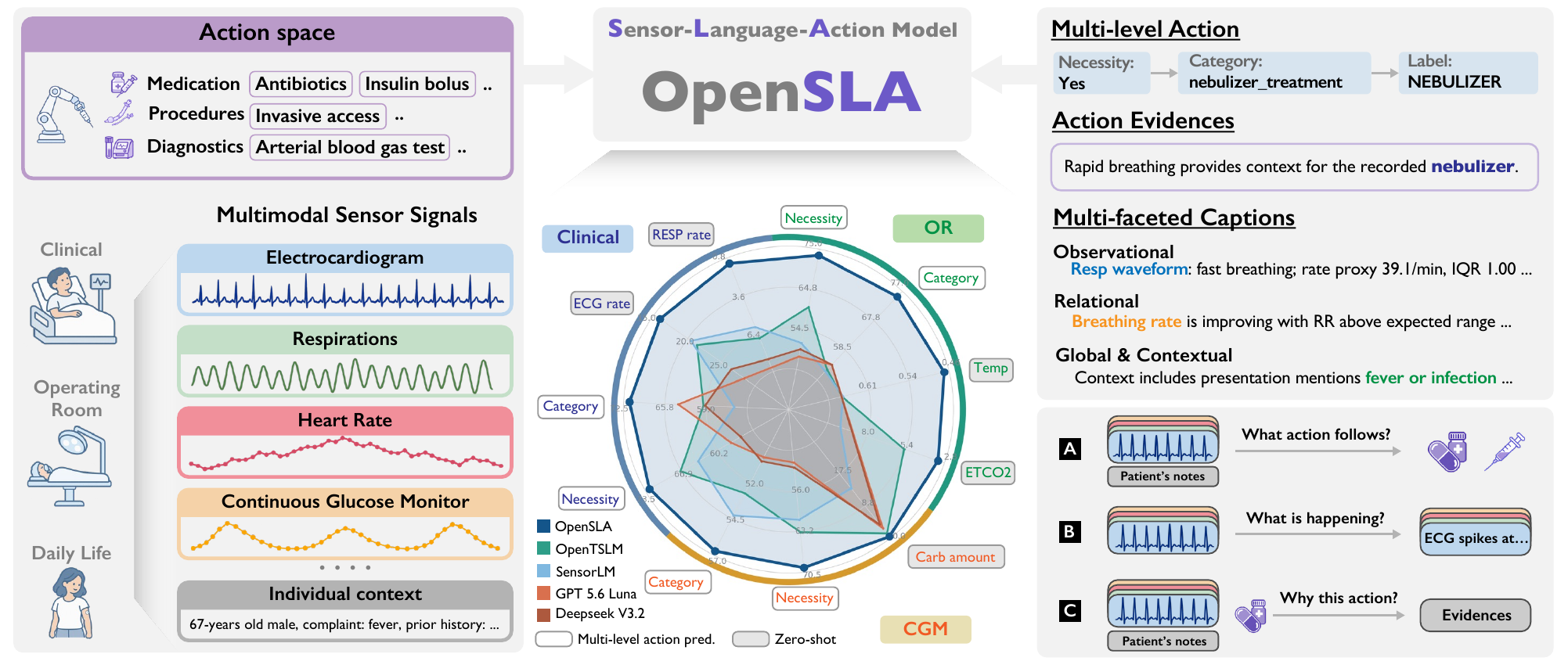}
    \vspace{-15pt}
    \caption{\small{\textbf{Sensor-Language-Action (SLA) Model.} We establish SLA as a unified paradigm for sensor-action modeling through a language interface over sensor observations, patient context, and actions. We instantiate SLA with \ours and evaluate it across more than {116,000} individuals, 79 sensor modalities, and \numaction action groups. In addition to the superior performance over state-of-the-art methods, \ours supports \textbf{(A)} multi-level action predictions, \textbf{(B)} sensor-state understanding, and \textbf{(C)} action-evidence grounding.}}
    \label{fig:overview}
\vspace{-5pt}
\end{figure}
\vspace{12pt}}

\newenvironment{Itemize}{
    \begin{itemize}[leftmargin=*]
    \setlength{\itemsep}{0pt}
    \setlength{\topsep}{0pt}
    \setlength{\partopsep}{0pt}
    \setlength{\parskip}{1pt}}
{\end{itemize}}
\setlength{\leftmargini}{9pt}

\vspace{-10pt}
\section{Introduction}
\label{sec:introduction}
\vspace{-5pt}

Intelligent systems do more than \textit{perceive}: they observe, understand, and \textit{\textbf{act}}.
The value of sensing, likewise, lies not only in describing what is happening, but in informing the actions that follow \citep{xu2026inertia, khasentino2025personal}.
For example, clinicians in operation rooms continuously integrate physiological signals with individual context, history, and prior interventions to decide whether to administer a medication, escalation, or continue observation \citep{PhysioNet-mimiciv-2.0}.

Yet, existing sensor models are centered on \textit{observing} and \textit{understanding}. Early methods map sensor signals to fixed labels or outcomes \citep{narayanswamy2025lsm,shuai2026osf}, while recent sensor-language models enable richer semantic descriptions and open-ended reasoning over sensor observations \citep{zhang2025sensorlm, langer2026opentslm, xu2026sleeplm}. However, these advances largely stop before \textbf{{action}}: downstream decisions are still modeled as separate prediction problems, each with its own formulation and action space \citep{lee2022development, weidman2025machine}. As a result, existing approaches provide no common mechanism for relating heterogeneous sensor evidence, contextual information, and the diverse actions they may support.

\textit{Natural language}, on the other hand, offers a natural bridge between sensing and action. Unlike fixed labels, language can jointly express task intent, individual context, physiological state, and action semantics, allowing heterogeneous sensor-action problems to be represented in a common form. This suggests a broader modeling paradigm in which sensor observations are grounded in \textit{language} and connected directly to \textit{actions}, rather than treating understanding and decision-making as separate stages. We therefore ask:
\vspace{-6pt}
\begin{center}
\emph{Can sensor data, language, and action be modeled through a unified framework?}
\end{center}

To fill the gap, we introduce \textbf{Sensor-Language-Action (SLA)} modeling, a unified framework that connects sensor observations, natural-language context, and actions through a common multimodal interface (see Fig. \ref{fig:overview}).
We further instantiate SLA with \ours, to our knowledge, the first general framework for jointly modeling sensor observations, language, and actions.
\ours extends sensor intelligence from understanding observations to reasoning about and explaining the actions they support, enabled by three key contributions:
\ding{182} A hierarchical, multi-faceted physiological captioning pipeline that aligns local sensor dynamics, cross-signal relations, individual context, and action-relevant evidence as language supervision;
\ding{183} The curation of a large-scale benchmark spanning more than 116,000 individuals, 79 sensor modalities, 3 healthcare domains, 5 tasks, and 60 action groups; and
\ding{184} a unified SLA architecture for hierarchical action prediction, state understanding, and action explanation that supports robust and scalable sensor-action modeling via language grounding.

To rigorously evaluate \ours, we benchmark it against state-of-the-art methods across six real-world cohorts and five core tasks spanning clinical care, operating rooms, and daily-life metabolic health. Extensive experiments demonstrate consistently strong performance in hierarchical action prediction, state understanding, and action explanation.
Further analyses reveal intriguing properties on action-evidence agreement, transfer to unseen action labels and external clinical cohorts, and future-state information in the learned representations.
Our contributions are as follows:
\vspace{-6pt}
\begin{Itemize}
    \item We formulate Sensor-Language-Action (SLA) modeling, a new framework that unifies sensor-action problems through multimodal language interface, and construct a large-scale benchmark spanning more than 116,000 individuals, 79 sensor modalities, 5 tasks, and 60 action groups.
    \vspace{2pt}
    \item We introduce a multi-faceted sensor captioning pipeline that jointly captures local sensor dynamics, cross-signal relations, individual-level context, and action-relevant evidence, providing language supervision tailored to sensor-action modeling.
    \vspace{2pt}
    \item We develop \ours, a unified SLA model that jointly supports hierarchical action prediction, state understanding, and action explanation, informed by systematic studies of how sensor signals, language, and actions should be modeled together.
    \vspace{2pt}
    \item We conduct extensive evaluations across diverse real-world tasks and verify the superior performance of \ours against SOTA methods.  Further analyses confirm its capabilities on language-guided evidence grounding and zero-shot generalization to unseen actions and cohorts.
\end{Itemize}

\vspace{-5pt}
\section{Related Work}
\label{sec:related-work}

\textbf{Sensor-Action Modeling.}
Predicting subsequent actions or decisions from observed sensor signals has been a long-standing problem \citep{lee2022development, weidman2025machine}. Existing approaches typically formulate this as supervised prediction, either learning direct mappings from sensor observations to action labels or incorporating additional textual context as conditioning information \citep{suresh2017clinical, hardan2024affordable, metwally2026insulin}. However, these approaches are less effective and flexible in incorporating heterogeneous individual context and other decision-relevant information. In contrast, we address this limitation by introducing \textit{language} as a flexible interface for sensor-action modeling.

\textbf{Sensor and Time-Series Foundation Models.}
Time-series foundation models \citep{ansari2025chronos2,woo2024moirai,goswami2024moment, yang2023simper} have shown strong performance across diverse sensor-centered tasks, including forecasting, classification, and representation learning. Beyond general-purpose time-series models, sensor-focused foundation models further specialize in physiological and wearable signals, covering ECG \citep{li2025electrocardiogram} and PPG \citep{pillai2025papagei}, continuous glucose monitoring \citep{lutsker2026foundation, li2026glucofm}, motion and wearable sensing \citep{xu2026inertia}, and sleep \citep{shuai2026osf}. Despite this progress, existing sensor foundation models largely focus on understanding states of input sensor signals, rather than translating sensor dynamics into action-relevant semantics. In contrast, \ours addresses this gap by jointly modeling sensor observations, language context, and actions through a shared sensor-language-action interface.

\begin{table*}[t]
\renewcommand{\arraystretch}{1.1}
\setlength{\tabcolsep}{6pt}
\centering
\caption{\small{
\textbf{Comparisons with representative sensor-language studies.}}}
\label{tab:related-work-data}
\vspace{-3pt}

\begin{adjustbox}{max width=\textwidth}
\footnotesize
\begin{tabular}{l ccc cccc cc}
\toprule[1.5pt]

\multirow{2.5}{*}{\textbf{Study}}
& \multicolumn{3}{c}{\textbf{Domain}}
& \multicolumn{4}{c}{\textbf{Data Scale}}
& \multicolumn{2}{c}{\textbf{Caption Content}} \\

\cmidrule(lr){2-4}
\cmidrule(lr){5-8}
\cmidrule(lr){9-10}

& Clinical
& OR
& Daily
& \multicolumn{1}{c}{\# Dataset}
& \multicolumn{1}{c}{\# Modality}
& \multicolumn{1}{c}{\# Individuals}
& \multicolumn{1}{c}{\# Action}
& Individual Context
& Action Evidence \\

\midrule
\citet{khasentino2025personal}
& \xmark
& \xmark
& \cmark
& 3
& 20
& 4.2k
& --
& \cmark
& \xmark \\

\citet{langer2026opentslm}
& \cmark
& \xmark
& \cmark
& 5
& 3
& N/R
& --
& \cmark
& \xmark \\


\citet{zhang2025sensorlm}
& \xmark
& \xmark
& \cmark
& 4
& 5
& 103k
& --
& \xmark
& \xmark \\

\citet{xu2026sleeplm}
& \cmark
& \xmark
& \xmark
& 5
& 12
& 10k
& --
& \xmark
& \xmark \\

\midrule

\rowcolor{gray!15}
\textbf{\ours (Ours)}
& \textbf{\cmark}
& \textbf{\cmark}
& \textbf{\cmark}
& \textbf{7}
& \textbf{\nummodality}
& \textbf{\numpatient}
& \textbf{\numaction}
& \textbf{\cmark}
& \textbf{\cmark} \\

\bottomrule[1.5pt]
\end{tabular}
\end{adjustbox}

\end{table*}

\textbf{Sensor-Language Modeling.} 
Language models have increasingly been applied to sensor analysis through direct prompting or dedicated sensor-language models \citep{shuai2026signal, li2026hearts, langer2026opentslm, xu2026sleeplm}. SensorLM \citep{zhang2025sensorlm} and SleepLM \citep{xu2026sleeplm} leverage paired sensor-text data and multi-level captioning to align sensor representations with language, demonstrating strong transferability across downstream tasks. OpenTSLM \citep{langer2026opentslm}, in contrast, combines time-series encoders with pretrained LLMs to support zero-shot language-based interaction with medical signals. Such models often rely on captioning pipelines for paired sensor-text supervision, but both their data and modeling pipelines are primarily designed for sensor-text alignment rather than action-relevant semantics. In this project, we introduce a multi-faceted captioning pipeline to capture a broader spectrum of information, forming the basis of our SLA model.

\vspace{-5pt}
\section{Sensor-Language-Action Dataset Construction}
\label{sec:task}
\label{sec:task:captioning}

\textbf{Data Regime.} 
We construct our sensor-action testbed from \numdataset development datasets and one external dataset spanning three real-world settings: clinical ICU prediction, operating room (OR) decision making, and daily-life continuous glucose monitoring (CGM). The clinical setting includes MC-MED \citep{kansal2025mcmed}, MIMIC-III \citep{johnson2016mimic}, and MIMIC-IV \citep{PhysioNet-mimiciv-2.0}; the OR setting includes MOVER \citep{samad2023mover} and VitalDB \citep{lee2022vitaldb}; and the CGM setting includes MetaboNet \citep{wolff2026metabonet} and PEDAP \citep{wadwa2023pedap}. We reserve MIMIC-IV as an external clinical cohort to evaluate cross-cohort generalization.
For each of the \numdataset development datasets, we construct individual-disjoint training, validation, and test splits, and combine the training sets within each setting.
The clinical and OR settings contain heterogeneous waveforms and numerical measurements, including ECG, plethysmography, respiration, blood pressure, SpO$_2$, and heart rate. The CGM domain contains glucose time series together with basal insulin delivery.

\textbf{Unified Multi-level Action Space.}
We formulate sensor-action modeling around a common question: \textit{given sensor observations and contextual information, what action should be taken?} In the clinical and OR settings, actions include medications, procedures, diagnostic evaluations, and treatment adjustments, whereas CGM actions correspond to insulin bolus delivery.
We organize actions across datasets into a shared three-level hierarchy: \ding{182} \textbf{action necessity}, indicating whether to act; \ding{183} \textbf{action category}, specifying the type of action; and \ding{184} \textbf{fine-grained action label}, identifying the specific action within a category. For example, administration of \texttt{Imipenem} can be represented as ``\emph{Yes}'' at the necessity level, ``\emph{antibiotics}'' at the category level, and ``\emph{Imipenem}'' at the fine-grained level. We derive action supervision from the recorded clinical decisions and interventions in each source dataset. Detailed dataset statistics and action definitions are provided in Appendix \ref{sec:appendix:dataset}.

\begin{figure}[t]
    \centering
    \includegraphics[width=\linewidth]{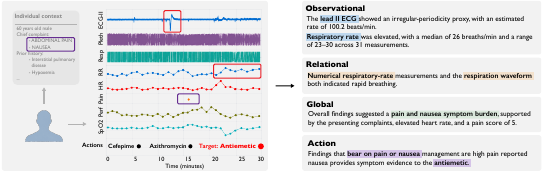}
    \vspace{-15pt}
    \caption{\small{\textbf{Multi-faceted sensor-action captioning pipeline.} Our captioning pipeline integrates four complementary facets of sensor-action information: (1) \textcolor{observation}{Observational} captions summarize signal-derived features; (2) \textcolor{relational}{Relational} captions describe cross-channel evidence; (3) \textcolor{global}{Global} captions capture patient-specific context; and (4) \textcolor{action}{Action} captions identify evidence from the preceding facets that is relevant to the action label.}}
    \label{fig:captioning-pipeline}
\vspace{-3pt}
\end{figure}

\textbf{Multi-faceted Captioning Pipeline.}
We aim to capture physiological evidence from sensor signals together with individual context relevant to decision making. Specifically, we structure each generated caption around four complementary facets: sensor observations, physiological relationships, individual-level context, and action-relevant evidence (Fig. \ref{fig:captioning-pipeline}). Implementation details are in Appendix \ref{sec:appendix:captioning}.

\textit{\textbf{\textcolor{observation}{Observational} Captions.}}
These captions describe channel-level physiological findings derived from established measurements and waveform characteristics (e.g., heart rate, respiratory rate, rhythm patterns). They translate directly observable sensor properties into explicit physiological descriptions.

\textit{\textbf{\textcolor{relational}{Relational} Captions.}}
We group physiologically related measurements into complementary feature sets and characterize their temporal and cross-channel relationships. For example, respiratory rate and breathing waveforms are jointly considered to characterize respiratory status, allowing multiple measurements to provide complementary evidence for the same underlying physiological process.

\textit{\textbf{\textcolor{global}{Global} Captions.}}
We integrate sensor-derived findings with individual-specific context, including available background information and prior user history, to summarize the overall physiological state. This places local sensor evidence within the broader context in which decisions are made.

\textit{\textbf{\textcolor{action}{Action} Captions.}}
We condition on recorded action categories to identify decision-relevant evidence from sensor measurements and individual context, and verbalize how this evidence supports the corresponding action. This formulation explicitly connects physiological observations with the evidence underlying downstream decisions.

Taken together, we construct a large-scale SLA testbed spanning \numdataset development datasets and the external MIMIC-IV cohort, with \nummodality sensor modalities, \numpatient individuals, and \numaction unique actions across three real-world settings. Our study substantially expands the scale and coverage of prior work (Table \ref{tab:related-work-data}), while additionally introducing structured action supervision.
We generate multiple caption combinations and evaluate their effectiveness in Sec. \ref{sec:experiments}. Unless otherwise specified, we use all four facets in our main experiments.
\section{Unifying Sensor-Action Tasks through a Language Interface}
\label{sec:method}

\begin{figure}[t]
    \centering
    \includegraphics[width=\linewidth]{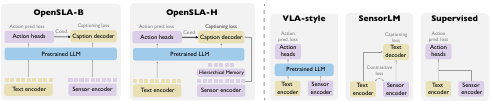}
    \vspace{-15pt}
    \caption{\small{\textbf{The \ours architecture and comparison with existing sensor modeling paradigms.} \ours integrates pretrained language knowledge with caption supervision for action-relevant sensor understanding and action-conditioned caption decoding for improved alignment between sensor, language, and actions.}}
    \label{fig:architecture}
\end{figure}

\subsection{Sensor-Language-Action Modeling}

We begin with a fundamental question: \textit{How can the knowledge encoded in language models be effectively used for sensor-action modeling?} 
To answer this, we study four language-model-centric paradigms: 
\ding{182} zero-shot inference with multimodal LLMs; \ding{183} VLA-style action-only fine-tuning \citep{kim2024openvla}; \ding{184} SensorLM-style sensor-language alignment \citep{zhang2025sensorlm}; and \ding{185} joint fine-tuning with action and language supervision.
These paradigms progressively couple language with sensor-action learning, from using pretrained multimodal knowledge without adaptation to directly aligning sensor observations, language, and actions during training (Fig. \ref{fig:architecture}).
We compare these paradigms on action prediction using VitalDB.
Joint action-language fine-tuning consistently performs best, outperforming VLA-style action-only training (Fig. \ref{fig:caption-variants-ablation}), while maintaining a substantial margin over SensorLM-style sensor-language alignment and zero-shot multimodal LLMs, including \gpt and \deepseek (Table \ref{tab:action-main-or}).
These results motivate the \textbf{Sensor-Language-Action (SLA)} paradigm, which jointly models sensor observations, natural-language instructions, and individual context, with both actions and language as prediction targets.

\begin{figure}[!htb]
    \centering
    \includegraphics[width=0.95\linewidth]{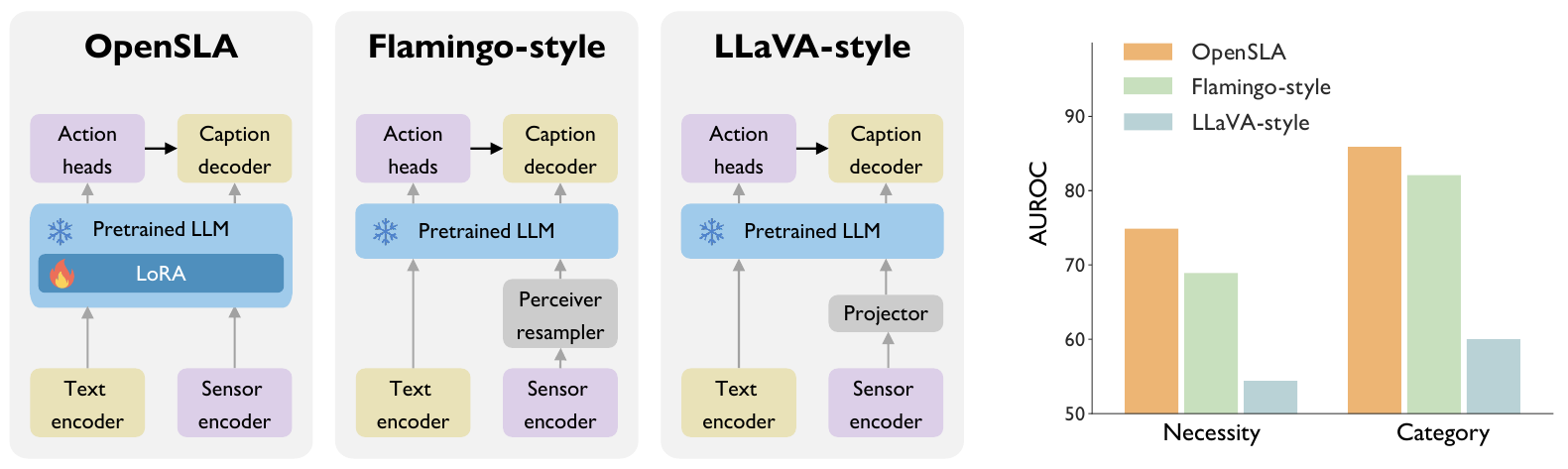}
    \vspace{-5pt}
    \caption{\textbf{Comparison of SLA alignment strategies.} \ours directly adapts the pretrained LLM with LoRA, outperforming Flamingo-style cross-attention and LLaVA-style projection.}
    \label{fig:multimodal-fusion}
\end{figure}

We next investigate how sensor representations should be integrated into the language backbone for SLA modeling. We consider three representative adaptation strategies, ranging from lightweight external alignment to direct adaptation of the LLM: \ding{182} LLaVA-style adapter training \citep{liu2023visual}; \ding{183} Flamingo-style cross-attention injection \citep{alayrac2022flamingo}; and \ding{184} direct Low-Rank Adaptation (LoRA) \citep{hu2021lora}. Fig. \ref{fig:multimodal-fusion} confirms direct LoRA adaptation achieves the strongest performance on most action-prediction tasks, outperforming both adapter-based and cross-attention-based alternatives. We therefore instantiate \ours using direct LoRA-based adaptation of the language backbone.

\subsection{\ours: Learning Sensor Dynamics through a Language Interface}

\ours comprises three main components: \ding{182} an LLM backbone with structured action-prediction heads; \ding{183} a hierarchical sensor encoder; and \ding{184} a multimodal fusion decoder (Fig. \ref{fig:architecture}). The model is jointly optimized for action prediction and autoregressive language generation, allowing action decisions and their sensor-grounded descriptions to be learned within a shared representation.

\textbf{Structured action prediction heads.} 
To support multi-level action prediction, we attach three prediction heads to the final-token representation of the LLM hidden states \(\mathbf H\), corresponding to action necessity, action category, and fine-grained action label. The necessity head uses a binary linear classifier to determine whether an action occurs, while the category head uses independent binary classifiers to predict applicable action categories. For fine-grained prediction, we use a similarity-based head that supports a flexible action space. Given an action category, textual descriptions of its candidate fine-grained actions are encoded into embeddings, and the action whose embedding is most similar to the LLM representation is selected.

\textbf{Hierarchical sensor encoder.} 
Sensor recordings can span long temporal windows and contain heterogeneous channel compositions. To efficiently represent these inputs, we design a hierarchical sensor encoder consisting of channel-wise signal encoding followed by a Hierarchical Memory module. Each channel is first independently processed by a frozen signal encoder, producing sensor tokens $\mathbf{Z}^{(0)}\in\mathbb{R}^{N_0\times D}$, where $N_0$ denotes the number of tokens and $D$ their feature dimension. The Hierarchical Memory then compresses these tokens across multiple resolutions while preserving both fine-grained sensor patterns and sample-level information.

The Hierarchical Memory contains two complementary pathways. The first recursively aggregates sensor tokens across multiple resolutions. Starting from $\mathbf{Z}^{(0)}$, we apply $L$ query layers, where the $l$-th layer uses $N_l$ learned queries to cross-attend to $\mathbf{Z}^{(l-1)}$, producing $\mathbf{Z}^{(l)} \in \mathbb{R}^{N_l \times D}$, with $N_l < N_{l-1}$. This recursive compression aggregates information over increasingly long temporal contexts. In parallel, a separate set of global queries attends directly to the original token sequence $\mathbf{Z}^{(0)}$ to capture sample-level semantics. The outputs of the two pathways are concatenated to form the sensor tokens provided to the language backbone.

\textbf{Multimodal fusion decoder.}
Although the hierarchical sensor encoder provides compact multi-resolution representations, the LLM hidden states may not retain the fine-grained sensor evidence required for grounded caption generation. We therefore introduce a gated residual caption decoder that explicitly conditions language generation on both predicted actions and sensor evidence. Specifically, two parallel branches cross-attend to the predicted action profile and retrieved sensor evidence, respectively. The action profile \(\mathbf P\) consists of an action-necessity token and action-category tokens, with each token combining the corresponding action embedding and predicted score. 
Conditioned on the LLM hidden states \(\mathbf H\) and action profile \(\mathbf P\), the sensor branch retrieves the most relevant tokens from the waveform and numerical sensor representations to form sensor evidence \(\mathbf S\).
The caption representation is then refined as
\begin{equation}
\mathbf{H}'
=
\mathbf{H}
\mathbin{+}
g_p f_p\bigl(\mathbf{H}, \mathbf{P}\bigr)
\mathbin{+}
g_s f_s\bigl(\mathbf{H}, \mathbf{S}\bigr),
\end{equation}
where $f_p$ and $f_s$ denote cross-attention modules over the action profile and sensor evidence, respectively, and $g_p$ and $g_s$ are learnable gates. The refined states \(\mathbf H'\) are passed through the original language-model head for autoregressive caption generation.

\textbf{Training objective.}
We denote the action prediction loss as $\mathcal{L}_{\mathrm{act}}$, which aggregates the losses for action necessity, category, and fine-grained action prediction, and optimize caption generation using autoregressive cross-entropy $\mathcal{L}_{\mathrm{text}}$. The overall training objective is $\mathcal{L}=\mathcal{L}_{\mathrm{act}}+\lambda_{\mathrm{text}}\mathcal{L}_{\mathrm{text}}$, where $\lambda_{\mathrm{text}}$ controls the contribution of language supervision. During training, we adapt the LLM backbone using LoRA and jointly optimize the hierarchical sensor encoder and multimodal fusion decoder, while keeping the remaining pretrained modules frozen. Implementation details are in Appendix \ref{sec:appendix:method:architecture}.

\vspace{-5pt}
\section{Experiment \& Results}
\label{sec:experiments}

\begin{table*}[!tb]
\renewcommand{\arraystretch}{1.1}
\setlength{\tabcolsep}{6pt}
\centering
\caption{\small{\textbf{Multi-level action prediction results for ICU clinical prediction.}}
}
\label{tab:action-main-clinical}
\vspace{-5pt}
\begin{adjustbox}{width=\textwidth}
\begin{tabular}{l *{5}{c} *{5}{c}}
\toprule[1.5pt]
\multirow{3.5}{*}{\textbf{Method}} & \multicolumn{5}{c}{\textbf{MC-MED}} & \multicolumn{5}{c}{\textbf{MIMIC-III}} \\
\cmidrule(lr){2-6}\cmidrule(lr){7-11}
 & \multicolumn{2}{c}{Necessity} & \multicolumn{2}{c}{Category} & \multicolumn{1}{c}{Fine-label} & \multicolumn{2}{c}{Necessity} & \multicolumn{2}{c}{Category} & \multicolumn{1}{c}{Fine-label} \\
\cmidrule(lr){2-3}\cmidrule(lr){4-5}\cmidrule(lr){6-6}\cmidrule(lr){7-8}\cmidrule(lr){9-10}\cmidrule(lr){11-11}
 & \multicolumn{1}{c}{AUROC$^\uparrow$} & \multicolumn{1}{c}{BAcc$^\uparrow$} & \multicolumn{1}{c}{AUROC$^\uparrow$} & \multicolumn{1}{c}{BAcc$^\uparrow$} & \multicolumn{1}{c}{BAcc$^\uparrow$} & \multicolumn{1}{c}{AUROC$^\uparrow$} & \multicolumn{1}{c}{BAcc$^\uparrow$} & \multicolumn{1}{c}{AUROC$^\uparrow$} & \multicolumn{1}{c}{BAcc$^\uparrow$} & \multicolumn{1}{c}{BAcc$^\uparrow$} \\
\midrule
\midrule
\multicolumn{11}{@{}l}{\textit{\textbf{Supervised:}}} \\
\textsc{ConvTrans}~\cite{foumani2024improving} & 65.2 \scriptsize$\pm1.8$ & 57.5 \scriptsize$\pm1.4$ & 59.8 \scriptsize$\pm1.8$ & 53.8 \scriptsize$\pm1.2$ & 54.7 \scriptsize$\pm2.1$ & 61.4 \scriptsize$\pm2.3$ & 59.7 \scriptsize$\pm1.8$ & 54.8 \scriptsize$\pm1.7$ & 51.7 \scriptsize$\pm0.8$ & 57.1 \scriptsize$\pm1.8$ \\
\textsc{MMIM}~\cite{han2021mmim} & 61.9 \scriptsize$\pm1.8$ & 53.4 \scriptsize$\pm1.2$ & {64.6} \scriptsize$\pm1.7$ & 56.0 \scriptsize$\pm1.0$ & 53.5 \scriptsize$\pm2.3$ & 66.0 \scriptsize$\pm2.1$ & 61.6 \scriptsize$\pm1.8$ & 64.5 \scriptsize$\pm1.6$ & 54.4 \scriptsize$\pm0.7$ & 60.2 \scriptsize$\pm1.8$ \\
\midrule
\multicolumn{11}{@{}l}{\textit{\textbf{TS Foundation Models:}}} \\
{Chronos-2}~\cite{ansari2025chronos2} & 62.5 \scriptsize$\pm1.8$ & 55.4 \scriptsize$\pm1.4$ & 56.9 \scriptsize$\pm1.9$ & 52.7 \scriptsize$\pm0.9$ & 51.8 \scriptsize$\pm2.6$ & 60.9 \scriptsize$\pm2.4$ & 56.6 \scriptsize$\pm1.8$ & 58.8 \scriptsize$\pm1.7$ & 52.5 \scriptsize$\pm0.8$ & 53.2 \scriptsize$\pm2.0$ \\
{Moirai}~\cite{woo2024moirai} & 62.5 \scriptsize$\pm1.8$ & 57.4 \scriptsize$\pm1.5$ & 54.4 \scriptsize$\pm2.1$ & 51.9 \scriptsize$\pm1.1$ & 51.7 \scriptsize$\pm2.4$ & 57.7 \scriptsize$\pm2.4$ & 55.7 \scriptsize$\pm1.8$ & 58.7 \scriptsize$\pm1.8$ & 52.8 \scriptsize$\pm0.8$ & 53.4 \scriptsize$\pm1.8$ \\
\midrule
\multicolumn{11}{@{}l}{\textit{\textbf{Multi-modal LLMs:}}} \\
{GPT-5.6-Luna}~\cite{openai2026gpt56luna} & -- & 60.6 \scriptsize$\pm1.6$ & -- & 63.1 \scriptsize$\pm1.3$ & 59.5 \scriptsize$\pm2.4$ & -- & 54.8 \scriptsize$\pm1.2$ & -- & 64.0 \scriptsize$\pm1.2$ & \textbf{79.5} \scriptsize$\pm1.6$ \\
{DeepSeek-V3.2}~\cite{deepseek2025v32} & -- & 56.1 \scriptsize$\pm1.8$ & -- & 58.2 \scriptsize$\pm1.3$ & 54.4 \scriptsize$\pm2.4$ & -- & 55.6 \scriptsize$\pm1.8$ & -- & 60.2 \scriptsize$\pm0.9$ & 72.2 \scriptsize$\pm2.0$ \\
\midrule
\multicolumn{11}{@{}l}{\textit{\textbf{TS-LLM:}}} \\
{OpenTSLM}~\cite{langer2026opentslm} & -- & 62.9 \scriptsize$\pm1.5$ & -- & 52.8 \scriptsize$\pm0.7$ & 50.1 \scriptsize$\pm1.1$ & -- & 71.2 \scriptsize$\pm1.7$ & -- & 66.0 \scriptsize$\pm1.3$ & 67.1 \scriptsize$\pm1.4$ \\
{SensorLM}~\cite{zhang2025sensorlm} & -- & 64.6 \scriptsize$\pm1.6$ & -- & 53.5 \scriptsize$\pm0.8$ & 50.4 \scriptsize$\pm1.2$ & -- & 62.9 \scriptsize$\pm1.8$ & -- & 55.3 \scriptsize$\pm0.7$ & 51.7 \scriptsize$\pm0.9$ \\
\midrule
\multicolumn{11}{@{}l}{\textit{\textbf{SLA:}}} \\
\rowcolor{gray!15}
\oursbase & \textbf{77.6} \scriptsize$\pm1.5$ & \textbf{72.5} \scriptsize$\pm1.5$ & \textbf{83.7} \scriptsize$\pm1.4$ & \underline{68.6} \scriptsize$\pm1.3$ & \underline{61.2} \scriptsize$\pm2.1$ & \textbf{84.0} \scriptsize$\pm1.6$ & \textbf{75.4} \scriptsize$\pm1.4$ & \textbf{88.3} \scriptsize$\pm1.5$ & \underline{73.1} \scriptsize$\pm1.6$ & 73.7 \scriptsize$\pm1.5$ \\
\rowcolor{gray!15}
\ourshq & \underline{77.0} \scriptsize$\pm1.5$ & \underline{72.2} \scriptsize$\pm1.4$ & \textbf{83.7} \scriptsize$\pm1.3$ & \textbf{68.7} \scriptsize$\pm1.3$ & \textbf{65.2} \scriptsize$\pm2.5$ & \underline{81.9} \scriptsize$\pm1.6$ & \underline{73.3} \scriptsize$\pm1.5$ & \underline{88.1} \scriptsize$\pm1.6$ & \textbf{74.5} \scriptsize$\pm1.7$ & \underline{76.5} \scriptsize$\pm1.2$ \\
\bottomrule[1.5pt]
\end{tabular}
\end{adjustbox}
\end{table*}

\textbf{Baselines.}
We compare \ours with four families of sensor modeling methods. \ding{182} \textbf{Supervised models.} We train ConvTrans \citep{foumani2024improving} and MMIM \citep{han2021mmim,hardan2024affordable} from scratch. \ding{183} \textbf{Time-series foundation models.} We linearly probe Chronos-2 \citep{ansari2025chronos2}, Moirai \citep{woo2024moirai}, Gluformer \citep{lutsker2026foundation} for action prediction. \ding{184} \textbf{Sensor-language models.} We adapt OpenTSLM \citep{langer2026opentslm} and SensorLM \citep{zhang2025sensorlm} using our training corpus. \ding{185} \textbf{Multimodal LLMs.} We prompt GPT-5.6-Luna \citep{openai2026gpt56luna} and DeepSeek-V3.2 \citep{deepseek2025v32} through CodeAct-style Python execution \citep{wang2024codeact}. Additional details are provided in Appendix \ref{sec:appendix:method:baselines}.

\textbf{Implementation details.}
We denote the base and hierarchical variants of our model by \oursbase and \ourshq, respectively. All \ours variants use a Qwen-3.5-2B backbone adapted with LoRA. For fair comparison, all methods trained on our cohorts use the same training data and compute budget, with signal encoders initialized from the same domain-specific DINO-style pretrained models \citep{oquab2023dinov2, shuai2026osf}, where applicable. Training is capped to 24 hours on H200 GPU nodes. Further details are provided in Appendix \ref{sec:appendix:method}.

\begin{table*}[!t]
\renewcommand{\arraystretch}{1.1}
\setlength{\tabcolsep}{6pt}
\centering
\caption{\small{\textbf{Multi-level action prediction results for operating room decision making.}}
}
\vspace{-5pt}
\label{tab:action-main-or}
\begin{adjustbox}{width=\textwidth}
\begin{tabular}{l *{5}{c} *{5}{c}}
\toprule[1.5pt]
\multirow{3.5}{*}{\textbf{Method}} & \multicolumn{5}{c}{\textbf{MOVER}} & \multicolumn{5}{c}{\textbf{VitalDB}} \\
\cmidrule(lr){2-6}\cmidrule(lr){7-11}
 & \multicolumn{2}{c}{Necessity} & \multicolumn{2}{c}{Category} & \multicolumn{1}{c}{Fine-label} & \multicolumn{2}{c}{Necessity} & \multicolumn{2}{c}{Category} & \multicolumn{1}{c}{Fine-label} \\
\cmidrule(lr){2-3}\cmidrule(lr){4-5}\cmidrule(lr){6-6}\cmidrule(lr){7-8}\cmidrule(lr){9-10}\cmidrule(lr){11-11}
 & \multicolumn{1}{c}{AUROC$^\uparrow$} & \multicolumn{1}{c}{BAcc$^\uparrow$} & \multicolumn{1}{c}{AUROC$^\uparrow$} & \multicolumn{1}{c}{BAcc$^\uparrow$} & \multicolumn{1}{c}{BAcc$^\uparrow$} & \multicolumn{1}{c}{AUROC$^\uparrow$} & \multicolumn{1}{c}{BAcc$^\uparrow$} & \multicolumn{1}{c}{AUROC$^\uparrow$} & \multicolumn{1}{c}{BAcc$^\uparrow$} & \multicolumn{1}{c}{BAcc$^\uparrow$} \\
\midrule
\midrule
\multicolumn{11}{@{}l}{\textit{\textbf{Supervised:}}} \\
\textsc{ConvTrans}~\cite{foumani2024improving} & 47.6 \scriptsize$\pm3.9$ & 48.0 \scriptsize$\pm3.3$ & 62.0 \scriptsize$\pm3.8$ & 56.9 \scriptsize$\pm3.3$ & 56.0 \scriptsize$\pm5.2$ & 52.1 \scriptsize$\pm3.4$ & 51.4 \scriptsize$\pm2.7$ & 70.9 \scriptsize$\pm4.2$ & 66.0 \scriptsize$\pm3.9$ & 51.4 \scriptsize$\pm12.6$ \\
\textsc{MMIM}~\cite{han2021mmim} & 53.5 \scriptsize$\pm3.6$ & 51.8 \scriptsize$\pm3.2$ & 63.2 \scriptsize$\pm3.6$ & 59.0 \scriptsize$\pm3.0$ & 55.9 \scriptsize$\pm4.0$ & 51.7 \scriptsize$\pm3.5$ & 51.1 \scriptsize$\pm2.5$ & 63.9 \scriptsize$\pm4.6$ & 56.8 \scriptsize$\pm4.1$ & 57.2 \scriptsize$\pm9.7$ \\
\midrule
\multicolumn{11}{@{}l}{\textit{\textbf{TS Foundation Models:}}} \\
{Chronos-2}~\cite{ansari2025chronos2} & 43.6 \scriptsize$\pm4.0$ & 46.3 \scriptsize$\pm3.0$ & 63.3 \scriptsize$\pm3.7$ & 60.5 \scriptsize$\pm3.3$ & 57.5 \scriptsize$\pm5.2$ & 53.9 \scriptsize$\pm3.4$ & 52.1 \scriptsize$\pm2.6$ & 65.7 \scriptsize$\pm4.0$ & 62.3 \scriptsize$\pm2.6$ & 58.0 \scriptsize$\pm12.0$ \\
{Moirai}~\cite{woo2024moirai} & 46.3 \scriptsize$\pm3.7$ & 45.7 \scriptsize$\pm2.9$ & 63.7 \scriptsize$\pm3.7$ & 57.0 \scriptsize$\pm3.1$ & 51.9 \scriptsize$\pm4.2$ & 49.8 \scriptsize$\pm3.7$ & 49.7 \scriptsize$\pm1.8$ & 60.9 \scriptsize$\pm5.8$ & 57.5 \scriptsize$\pm4.0$ & 54.8 \scriptsize$\pm16.2$ \\
\midrule
\multicolumn{11}{@{}l}{\textit{\textbf{Multi-modal LLMs:}}} \\
{GPT-5.6-Luna}~\cite{openai2026gpt56luna} & -- & 45.0 \scriptsize$\pm3.3$ & -- & 52.4 \scriptsize$\pm1.3$ & \textbf{69.2} \scriptsize$\pm4.6$ & -- & 50.1 \scriptsize$\pm2.3$ & -- & 55.7 \scriptsize$\pm2.3$ & \underline{78.6} \scriptsize$\pm1.6$ \\
{DeepSeek-V3.2}~\cite{deepseek2025v32} & -- & 48.1 \scriptsize$\pm3.2$ & -- & 54.3 \scriptsize$\pm2.4$ & 55.9 \scriptsize$\pm4.0$ & -- & 50.8 \scriptsize$\pm1.9$ & -- & 53.8 \scriptsize$\pm1.8$ & 66.2 \scriptsize$\pm3.4$ \\
\midrule
\multicolumn{11}{@{}l}{\textit{\textbf{TS-LLM:}}} \\
{OpenTSLM}~\cite{langer2026opentslm} & -- & 71.5 \scriptsize$\pm3.0$ & -- & 55.2 \scriptsize$\pm1.9$ & 52.3 \scriptsize$\pm1.9$ & -- & 48.9 \scriptsize$\pm2.2$ & -- & 49.9 \scriptsize$\pm1.0$ & 51.2 \scriptsize$\pm2.5$ \\
{SensorLM}~\cite{zhang2025sensorlm} & -- & 49.9 \scriptsize$\pm1.0$ & -- & 52.9 \scriptsize$\pm1.9$ & 50.1 \scriptsize$\pm0.2$ & -- & 52.1 \scriptsize$\pm1.9$ & -- & 51.0 \scriptsize$\pm0.6$ & 51.7 \scriptsize$\pm1.0$ \\
\midrule
\multicolumn{11}{@{}l}{\textit{\textbf{SLA:}}} \\
\rowcolor{gray!15}
\oursbase & \underline{83.4} \scriptsize$\pm3.1$ & \underline{74.2} \scriptsize$\pm3.0$ & \textbf{80.3} \scriptsize$\pm2.7$ & \underline{69.0} \scriptsize$\pm2.7$ & 66.4 \scriptsize$\pm3.2$ & \underline{74.8} \scriptsize$\pm2.8$ & \underline{59.0} \scriptsize$\pm1.7$ & \underline{85.9} \scriptsize$\pm1.5$ & \underline{79.5} \scriptsize$\pm2.1$ & \textbf{78.9} \scriptsize$\pm1.0$ \\
\rowcolor{gray!15}
\ourshq & \textbf{85.9} \scriptsize$\pm2.7$ & \textbf{80.6} \scriptsize$\pm2.8$ & \underline{79.9} \scriptsize$\pm3.1$ & \textbf{69.9} \scriptsize$\pm2.3$ & \underline{67.6} \scriptsize$\pm2.6$ & \textbf{77.1} \scriptsize$\pm2.4$ & \textbf{66.3} \scriptsize$\pm2.4$ & \textbf{86.3} \scriptsize$\pm1.4$ & \textbf{80.8} \scriptsize$\pm1.8$ & 78.5 \scriptsize$\pm1.3$ \\
\bottomrule[1.5pt]
\end{tabular}
\end{adjustbox}
\end{table*}

\begin{table*}[!t]
\renewcommand{\arraystretch}{1.1}
\setlength{\tabcolsep}{6pt}
\centering
\caption{\small{\textbf{Multi-level action prediction results for CGM-based metabolic health prediction.}}
}
\vspace{-5pt}
\label{tab:action-main-cgm}
\begin{adjustbox}{width=0.85\textwidth}
\begin{tabular}{l *{4}{c} *{4}{c}}
\toprule[1.5pt]
\multirow{3.5}{*}{\textbf{Method}} & \multicolumn{4}{c}{\textbf{MetaboNet}} & \multicolumn{4}{c}{\textbf{PEDAP}} \\
\cmidrule(lr){2-5}\cmidrule(lr){6-9}
 & \multicolumn{2}{c}{Necessity} & \multicolumn{2}{c}{Category} & \multicolumn{2}{c}{Necessity} & \multicolumn{2}{c}{Category} \\
\cmidrule(lr){2-3}\cmidrule(lr){4-5}\cmidrule(lr){6-7}\cmidrule(lr){8-9}
 & \multicolumn{1}{c}{AUROC$^\uparrow$} & \multicolumn{1}{c}{BAcc$^\uparrow$} & \multicolumn{1}{c}{AUROC$^\uparrow$} & \multicolumn{1}{c}{BAcc$^\uparrow$} & \multicolumn{1}{c}{AUROC$^\uparrow$} & \multicolumn{1}{c}{BAcc$^\uparrow$} & \multicolumn{1}{c}{AUROC$^\uparrow$} & \multicolumn{1}{c}{BAcc$^\uparrow$} \\
\midrule
\midrule
\multicolumn{9}{@{}l}{\textit{\textbf{Supervised:}}} \\
\textsc{ConvTrans}~\cite{foumani2024improving} & 66.5 \scriptsize$\pm1.5$ & 62.2 \scriptsize$\pm1.3$ & 53.8 \scriptsize$\pm0.9$ & 52.7 \scriptsize$\pm0.7$ & 79.0 \scriptsize$\pm2.8$ & 72.8 \scriptsize$\pm2.6$ & \textbf{70.5} \scriptsize$\pm2.8$ & \textbf{66.2} \scriptsize$\pm2.5$ \\
\textsc{MMIM}~\cite{han2021mmim} & 66.3 \scriptsize$\pm1.9$ & 60.2 \scriptsize$\pm1.6$ & 62.7 \scriptsize$\pm2.1$ & 58.7 \scriptsize$\pm1.6$ & 64.7 \scriptsize$\pm3.5$ & 61.7 \scriptsize$\pm3.6$ & 58.5 \scriptsize$\pm3.6$ & \underline{56.3} \scriptsize$\pm3.0$ \\
\midrule
\multicolumn{9}{@{}l}{\textit{\textbf{TS Foundation Models:}}} \\
{Chronos-2}~\cite{ansari2025chronos2} & 63.6 \scriptsize$\pm1.5$ & 60.3 \scriptsize$\pm1.3$ & 52.5 \scriptsize$\pm0.7$ & 51.7 \scriptsize$\pm0.5$ & 71.7 \scriptsize$\pm3.2$ & 66.1 \scriptsize$\pm2.6$ & 55.8 \scriptsize$\pm1.9$ & 54.0 \scriptsize$\pm1.5$ \\
{GluFormer}~\cite{lutsker2026foundation} & 52.0 \scriptsize$\pm1.1$ & 51.3 \scriptsize$\pm0.8$ & 51.0 \scriptsize$\pm1.2$ & 50.5 \scriptsize$\pm0.9$ & 53.5 \scriptsize$\pm1.3$ & 52.2 \scriptsize$\pm1.2$ & 53.3 \scriptsize$\pm2.7$ & 52.8 \scriptsize$\pm1.6$ \\
\midrule
\multicolumn{9}{@{}l}{\textit{\textbf{Multi-modal LLMs:}}} \\
{GPT-5.6-Luna}~\cite{openai2026gpt56luna} & -- & 49.9 \scriptsize$\pm0.4$ & -- & 50.4 \scriptsize$\pm0.2$ & -- & 52.0 \scriptsize$\pm0.8$ & -- & 50.1 \scriptsize$\pm0.3$ \\
{DeepSeek-V3.2}~\cite{deepseek2025v32} & -- & 51.2 \scriptsize$\pm0.7$ & -- & 50.5 \scriptsize$\pm0.3$ & -- & 52.5 \scriptsize$\pm0.5$ & -- & 50.6 \scriptsize$\pm0.3$ \\
\midrule
\multicolumn{9}{@{}l}{\textit{\textbf{TS-LLM:}}} \\
{OpenTSLM}~\cite{langer2026opentslm} & -- & 62.1 \scriptsize$\pm1.5$ & -- & 53.9 \scriptsize$\pm1.1$ & -- & 64.6 \scriptsize$\pm3.7$ & -- & 51.6 \scriptsize$\pm1.2$ \\
{SensorLM}~\cite{zhang2025sensorlm} & -- & 62.9 \scriptsize$\pm1.5$ & -- & 56.8 \scriptsize$\pm1.4$ & -- & 59.4 \scriptsize$\pm3.0$ & -- & 51.8 \scriptsize$\pm0.5$ \\
\midrule
\multicolumn{9}{@{}l}{\textit{\textbf{SLA:}}} \\
\rowcolor{gray!15}
\oursbase & \textbf{73.3} \scriptsize$\pm1.5$ & \textbf{65.0} \scriptsize$\pm1.3$ & \underline{66.5} \scriptsize$\pm2.1$ & \underline{59.2} \scriptsize$\pm2.2$ & \textbf{81.2} \scriptsize$\pm3.3$ & \textbf{75.5} \scriptsize$\pm2.9$ & \underline{65.2} \scriptsize$\pm5.5$ & \underline{56.3} \scriptsize$\pm1.9$ \\
\rowcolor{gray!15}
\ourshq & \underline{71.9} \scriptsize$\pm1.5$ & \underline{64.3} \scriptsize$\pm1.3$ & \textbf{67.2} \scriptsize$\pm2.2$ & \textbf{59.6} \scriptsize$\pm2.1$ & \underline{80.9} \scriptsize$\pm3.3$ & \underline{75.1} \scriptsize$\pm2.8$ & 59.7 \scriptsize$\pm2.4$ & 53.9 \scriptsize$\pm0.8$ \\
\bottomrule[1.5pt]
\end{tabular}
\end{adjustbox}
\end{table*}

\textbf{Evaluation.}
We evaluate \ours against the baselines across the clinical, OR, and CGM domains on five tasks spanning three capabilities. 
\ding{182} \textbf{Action Prediction.} We report macro-AUROC and balanced accuracy for action necessity and category prediction, and balanced accuracy for fine-grained action prediction. Metrics are macro-averaged across labels at each action granularity. Fine-grained performance is evaluated within ground-truth-positive action categories. 
\ding{183} \textbf{State Understanding.} Following \citep{xu2026sleeplm}, we report the mean absolute error (MAE) between sensor measurements extracted from generated captions and their corresponding ground-truth values. 
\ding{184} \textbf{Action Explanation.} Following prior sensor-language evaluation \citep{zhang2025sensorlm,xu2026sleeplm}, we measure reference-based agreement between generated explanations and action-evidence captions using BERTScore \citep{zhang2019bertscore}. Scores are computed with DistilBERT \citep{sanh2019distilbert} following the setup of \citep{xu2025radeval}.

\textbf{\ours predicts actions across domains and granularities.} 
We evaluate action necessity, category, and fine-grained action prediction across the three domains, where applicable. As shown in Table \ref{tab:action-main-clinical}, Table \ref{tab:action-main-or}, and Table \ref{tab:action-main-cgm}, \ours consistently achieves strong performance across domains and action granularities, with particularly large gains for action necessity and category prediction. The results also reveal distinct domain-specific patterns. Sensor-only action prediction is most challenging in the OR setting, while \ours achieves its largest gains in the clinical and OR domains, suggesting that individual-level textual context is especially informative for context-dependent clinical decisions. Across most settings, performance improves from necessity prediction to category and fine-grained prediction, as the candidate action space becomes increasingly constrained once an intervention and its category are known. CGM is a notable exception: action-category prediction remains challenging because insulin decisions can depend on future carbohydrate intake that is unavailable in the observed context.

\begin{table*}[!t]
\renewcommand{\arraystretch}{1.1}
\setlength{\tabcolsep}{6pt}
\centering
\setlength{\aboverulesep}{1pt}
\setlength{\belowrulesep}{1.5pt}
\caption{\small{\textbf{Caption-based sensor-state estimation results.} Results are evaluated using MAE$^\downarrow$ across sensor-state estimation tasks.}}
\label{tab:state-understanding-main}
\vspace{-5pt}
\begin{adjustbox}{width=1\textwidth}
\begin{tabular}{l *{3}{c} *{3}{c} *{2}{c}}
\toprule[1.5pt]
\multirow{3}{*}{\textbf{Method}}
& \multicolumn{3}{c}{\textbf{Clinical}}
& \multicolumn{3}{c}{\textbf{Operating Room}}
& \multicolumn{2}{c}{\textbf{CGM}} \\
\cmidrule(lr){2-4}
\cmidrule(lr){5-7}
\cmidrule(lr){8-9}

& \multicolumn{1}{c}{\shortstack{Cardiac Index}}
& \multicolumn{1}{c}{\shortstack{ECG-II Rate}}
& \multicolumn{1}{c}{\shortstack{RESP Rate}}
& \multicolumn{1}{c}{\shortstack{Resp. Rate}}
& \multicolumn{1}{c}{\shortstack{Temperature}}
& \multicolumn{1}{c}{ETCO$_2$}
& \multicolumn{1}{c}{\shortstack{Basal Rate}}
& \multicolumn{1}{c}{\shortstack{Carb Amount}} \\
\midrule
\midrule

\multicolumn{9}{@{}l}{\textit{\textbf{Multi-modal LLMs: }}} \\
{GPT-5.6-Luna}~\cite{openai2026gpt56luna}
& -- & 28.4 & --
& -- & -- & --
& \textbf{0.07} & 2.70 \\
{DeepSeek-V3.2}~\cite{deepseek2025v32}
& -- & 26.5 & 8.27
& -- & -- & --
& 0.43 & 3.70 \\

\midrule
\multicolumn{9}{@{}l}{\textit{\textbf{TS-LLM:}}} \\
{OpenTSLM}~\cite{langer2026opentslm}
& 1.13 & 21.4 & 6.71
& 2.66 & 0.66 & 5.47
& 0.58 & 1.20 \\
{SensorLM}~\cite{zhang2025sensorlm}
& -- & 20.5 & 5.87
& 2.56 & 0.66 & 9.76
& 0.54 & 13.5 \\

\midrule
\multicolumn{9}{@{}l}{\textit{\textbf{SLA:}}} \\
\rowcolor{gray!15}
\oursbase
& \textbf{0.93} & \underline{20.1} & \underline{1.59}
& \underline{2.48} & \underline{0.64} & \underline{5.13}
& \textbf{0.07} & \underline{0.73} \\
\rowcolor{gray!15}
\ourshq
& \underline{1.01} & \textbf{15.9} & \textbf{1.17}
& \textbf{1.41} & \textbf{0.47} & \textbf{3.18}
& \underline{0.35} & \textbf{0.38} \\

\bottomrule[1.5pt]
\end{tabular}
\end{adjustbox}
\vspace{8pt}
\end{table*}

\begin{figure}[!t]
    \centering
    \includegraphics[width=1\linewidth]{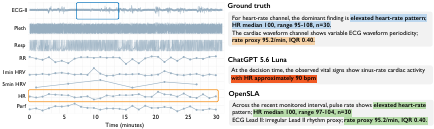}
    \vspace{-15pt}
    \caption{\small{\textbf{Sensor-grounded caption generation.} \ours captures both sensor states and fine-grained signal statistics, while the LLM baseline misses key physiological details.}}
    \label{fig:evidence grounding}
\end{figure}

\textbf{Language supervision enriches sensor-action representations beyond action prediction.}
We further assess whether the generated captions accurately recover physiological quantities from the observed sensor history, without training an additional regression head. As shown in Table \ref{tab:state-understanding-main}, \ours yields the most accurate physiological state estimates across the compared methods, indicating that its generated language retains detailed information about the underlying sensor state. The gains are most pronounced in the clinical setting, where effective prediction requires integrating complementary information across multiple sensor modalities.

\begin{table*}[!t]
\renewcommand{\arraystretch}{1.1}
\setlength{\tabcolsep}{6pt}
\centering
\setlength{\aboverulesep}{1pt}
\setlength{\belowrulesep}{1.5pt}
\caption{\small{\textbf{Action explanation results.} BERTScore$^\uparrow$~\cite{zhang2019bertscore} (F1) measures semantic agreement between generated text and reference action-evidence captions.}}
\vspace{-5pt}
\label{tab:action-explanation-main}
\begin{adjustbox}{max width=0.7\textwidth}
\begin{tabular}{l *{2}{c} *{2}{c} *{2}{c}}
\toprule[1.5pt]
\multirow{3}{*}{\textbf{Method}}
& \multicolumn{2}{c}{\textbf{Clinical}}
& \multicolumn{2}{c}{\textbf{Operating Room}}
& \multicolumn{2}{c}{\textbf{CGM}} \\
\cmidrule(lr){2-3}\cmidrule(lr){4-5}\cmidrule(lr){6-7}
& MC-MED & MIMIC-III & MOVER & VitalDB & MetaboNet & PEDAP \\
\midrule
\midrule
\multicolumn{7}{@{}l}{\textit{\textbf{Multi-modal LLMs:}}} \\
{GPT-5.6-Luna}~\cite{openai2026gpt56luna} & 75.1 & 75.1 & 70.7 & 71.4 & 76.1 & 76.6 \\
{DeepSeek-V3.2}~\cite{deepseek2025v32} & 73.8 & 73.2 & 69.9 & 71.2 & 74.5 & 75.5 \\
\midrule
\multicolumn{7}{@{}l}{\textit{\textbf{TS-LLM:}}} \\
{OpenTSLM}~\cite{langer2026opentslm} & 79.5 & \underline{80.3} & \textbf{81.7} & \underline{80.7} & \underline{84.4} & \underline{83.2} \\
{SensorLM}~\cite{zhang2025sensorlm} & 77.2 & 78.8 & 67.9 & 69.8 & 76.1 & 76.4 \\
\midrule
\multicolumn{7}{@{}l}{\textit{\textbf{SLA:}}} \\
\rowcolor{gray!15}
\oursbase & \underline{81.5} & 78.6 & \underline{81.5} & \textbf{80.9} & \textbf{85.6} & \textbf{87.6} \\
\rowcolor{gray!15}
\ourshq & \textbf{82.0} & \textbf{80.9} & 81.0 & \underline{80.7} & 80.4 & 76.7 \\
\bottomrule[1.5pt]
\end{tabular}
\end{adjustbox}
\vspace{10pt}
\end{table*}

\textbf{\ours captures action-relevant evidence through language.}
Beyond structured action prediction, we assess whether generated descriptions recover the evidence associated with each recorded action. Table \ref{tab:action-explanation-main} confirms that \ours achieves the strongest overall agreement with reference action-evidence captions across domains. Overall, \ours more faithfully captures the sensor and contextual evidence associated with downstream actions.

To examine this grounding directly, we qualitatively compare generated captions with their underlying sensor observations on a representative case. As shown in Fig. \ref{fig:evidence grounding}, \ours produces more precise and physiologically grounded descriptions than zero-shot multimodal LLMs, including \gpt and \deepseek, with closer agreement between the generated explanation and the observed sensor state.
\vspace{-3pt}
\section{Generalization and Predictive Structure in SLA Representations}
\vspace{-2pt}

\begin{figure}[!t]
    \centering
    \includegraphics[width=0.9\linewidth]{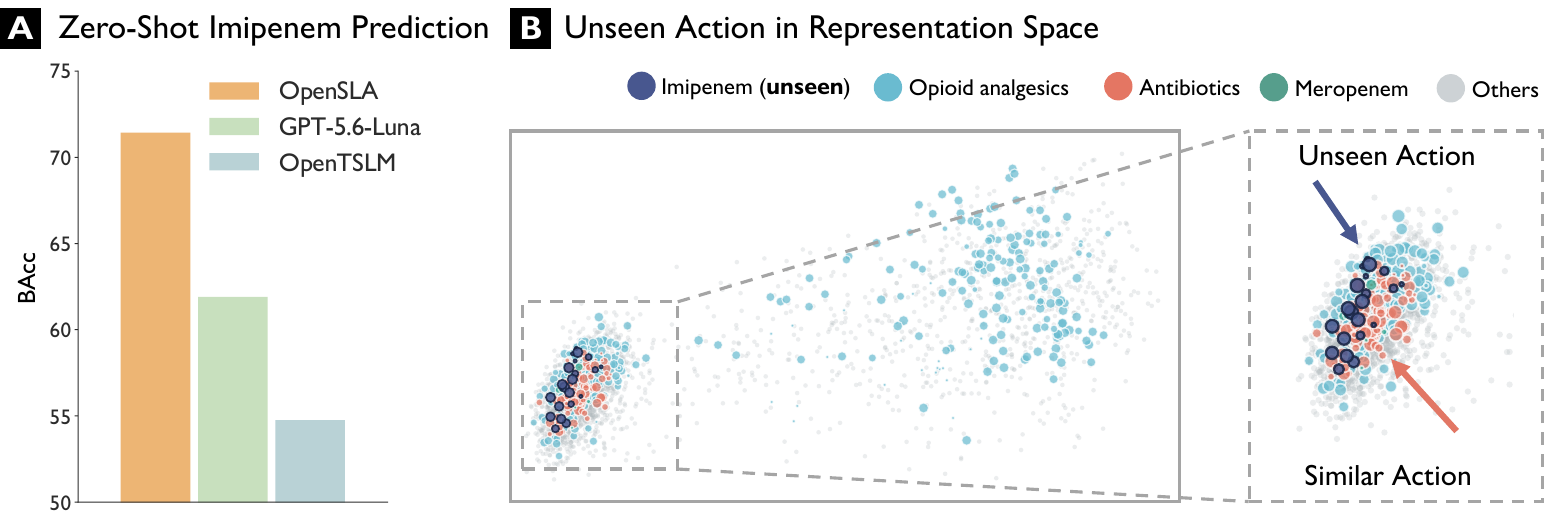}
    \vspace{-5pt}
    \caption{\small{
\textbf{Generalization to unseen actions.}
\textbf{(A)} Zero-shot prediction of the held-out action \texttt{Imipenem}. \textbf{(B)} Visualization of the learned \ours action representation space, where the unseen \texttt{Imipenem} representation lies near semantically related antibiotics.}}
    \label{fig:embedding}
\end{figure}

\textbf{Generalization to unseen actions.}
Beyond in-distribution action prediction, we ask whether sensor-language-action alignment induces a representation space that supports generalization to previously unseen actions. We first examine the geometry of representations learned on MIMIC-III by projecting backbone features into a PCA-reduced space. 
As shown in Fig. \ref{fig:embedding}, semantically related actions occupy nearby regions. For example, \texttt{antibiotic medications} form a coherent cluster, and an unseen action \texttt{Imipenem} (excluded from training but belonging to the same semantic category), is positioned near other antibiotics.

Motivated by this structure, we test whether \ours can predict unseen actions directly from the learned representation space. We hold out \texttt{Imipenem} during training and perform zero-shot prediction by matching sample representations to textual action embeddings through similarity. \ours achieves the highest balanced accuracy on this unseen-action task (Fig. \ref{fig:embedding}), suggesting that SLA alignment supports transfer through the semantic structure shared across related actions.

\begin{figure}[!t]
    \centering
    \includegraphics[width=\linewidth]{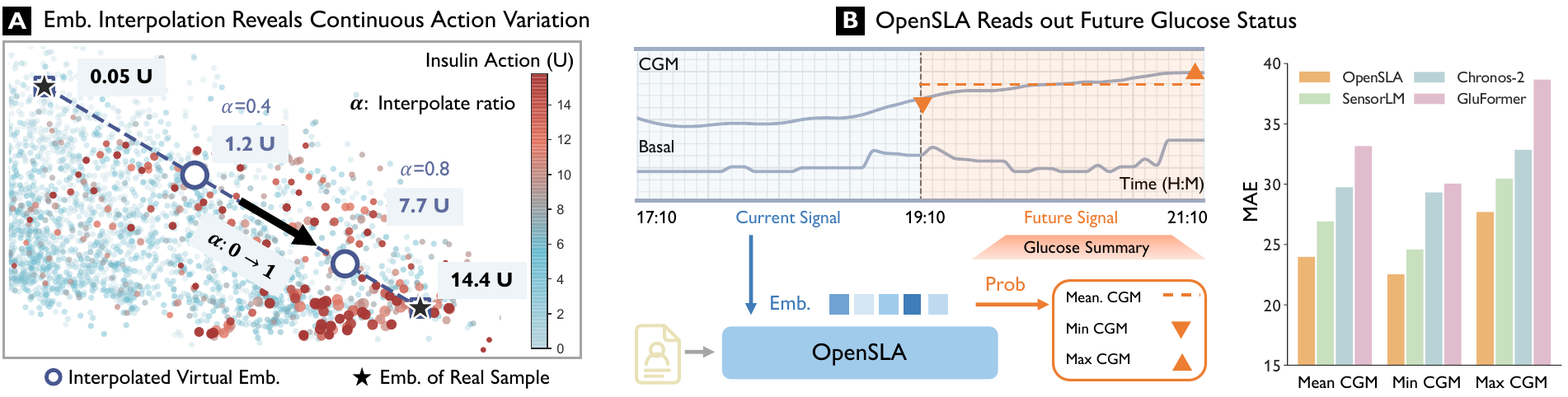}
    \vspace{-15pt}
    \caption{\small{
\textbf{Continuous action structure and future-state information emerge in SLA representations.}
\textbf{(A)} Interpolating between learned representations yields smoothly varying insulin actions among the nearest real samples, revealing continuous action structure in the embedding space. \textbf{(B)} Representations derived from current CGM observations support accurate readout of future glucose statistics, indicating that SLA representations retain information predictive of subsequent sensor states.}}
    \label{fig:world modeling}
\end{figure}

\textbf{Continuous action structure emerges in the learned representations.}
We next ask whether this organization extends beyond discrete action categories to continuous action spaces. We interpolate between representations associated with two distinct insulin actions. At each interpolated point, we retrieve the nearest real-sample representation and examine its corresponding action value.  Fig. \ref{fig:world modeling} verifies the retrieved insulin actions vary monotonically along the interpolation trajectory, revealing a continuous action structure within the learned representation space.

\textbf{Future sensor states are recoverable from SLA representations.}
We further ask whether representations learned without explicit future-state supervision retain information about subsequent sensor dynamics. For each 2-hour CGM observation window, we define future targets using the mean, minimum, maximum, and slope of glucose over the subsequent 2 hours. With the SLA backbone frozen, we train ridge readouts on these targets using individual-grouped nested cross-validation. Fig. \ref{fig:world modeling} confirms that \ours representations enable accurate prediction of future glucose states. These results suggest that SLA learning captures predictive physiological structure beyond the action targets used during backbone training.

\begin{table}[ht]
\centering
\begin{minipage}[t]{0.47\linewidth}
\vspace{0pt}
\noindent\textbf{Transfer to unseen datasets.}
To examine generalization beyond the training cohorts, we directly apply \ours to the external MIMIC-IV cohort without adaptation. Table \ref{tab:appendix:mimiciv-action} shows that \ours outperforms baselines on both action necessity and category prediction. These results demonstrate that the learned sensor-language-action associations transfer to an unseen clinical cohort under a shared action vocabulary.
\end{minipage}\hfill
\begin{minipage}[t]{0.50\linewidth}
\vspace{0pt}
\begin{minipage}[t]{\linewidth}
\renewcommand{\arraystretch}{1.1}
\setlength{\tabcolsep}{8pt}
\centering
\captionof{table}{\small{\textbf{Zero-shot transfer to the external MIMIC-IV cohort.}}}
\vspace{-5pt}
\label{tab:appendix:mimiciv-action}
\small
\begin{adjustbox}{max width=\linewidth}
\begin{tabular}{l cc cc}
\toprule[1.5pt]
\multirow{2}{*}{\textbf{Method}} & \multicolumn{2}{c}{\textbf{Necessity}} & \multicolumn{2}{c}{\textbf{Category}} \\
\cmidrule(lr){2-3}\cmidrule(lr){4-5}
& AUROC$^\uparrow$ & BAcc$^\uparrow$ & AUROC$^\uparrow$ & BAcc$^\uparrow$ \\
\midrule
{OpenTSLM}~\cite{langer2026opentslm} & -- & 69.30 & -- & 54.33 \\
{SensorLM}~\cite{zhang2025sensorlm} & -- & 67.70 & -- & 50.02 \\
\midrule
\rowcolor{gray!15}
\oursbase & \textbf{81.75} & \textbf{74.40} & \underline{74.64} & \underline{56.35} \\
\rowcolor{gray!15}
\ourshq & \underline{80.52} & \underline{73.00} & \textbf{74.90} & \textbf{58.49} \\
\bottomrule[1.5pt]
\end{tabular}
\end{adjustbox}
\end{minipage}

\end{minipage}
\end{table}

\begin{figure}[!t]
\input{figures/arxiv/7_ablation}
\end{figure}

\textbf{Which model components support action prediction?}
We ablate the main components of \ours to identify their contributions to sensor-action modeling. As shown in Fig. \ref{fig:caption-variants-ablation}, the full model achieves the strongest overall performance. The largest degradation occurs under VLA-style action-only training, highlighting the benefit of introducing language supervision into sensor-action learning. Removing action conditioning or the multimodal fusion decoder further reduces performance, indicating that explicitly connecting action predictions with sensor-grounded language provides additional gains.

\begin{wraptable}{r}{0.55\linewidth}
\vspace{-9pt}
\renewcommand{\arraystretch}{1.1}
\setlength{\tabcolsep}{4pt}
\centering

\caption{\small{\textbf{Ablation of caption supervision facets.}
AUROC$^\uparrow$ (\%) is used as the metric.}}
\label{tab:action-ablation-or-caption-types}

\vspace{-6pt}
\small

\begin{adjustbox}{max width=\linewidth}
\begin{tabular}{cccc cc cc}
\toprule[1.5pt]
\multicolumn{4}{c}{\textbf{Caption Variants}} 
& \multicolumn{2}{c}{\textbf{MOVER}} 
& \multicolumn{2}{c}{\textbf{VitalDB}} \\
\cmidrule(lr){1-4}\cmidrule(lr){5-6}\cmidrule(lr){7-8}
Obs. & Rel. & Global & Action & Necessity & Category & Necessity & Category \\
\midrule
\cmark & \xmark & \xmark & \xmark & 81.5 & 76.2 & 64.1 & 81.1 \\
\xmark & \xmark & \xmark & \cmark & 80.0 & 73.3 & 60.9 & 80.2 \\
\cmark & \xmark & \xmark & \cmark & 81.0 & 76.1 & 65.2 & 82.7 \\
\cmark & \cmark & \xmark & \xmark & 79.2 & 76.4 & 65.9 & 82.6 \\
\cmark & \xmark & \cmark & \cmark & \underline{81.6} & 77.3 & \underline{66.8} & \underline{83.9} \\
\cmark & \cmark & \xmark & \cmark & 80.7 & 76.4 & 64.4 & 82.2 \\
\cmark & \cmark & \cmark & \xmark & \underline{81.6} & \underline{77.5} & 64.3 & 82.3 \\
\rowcolor{gray!15}
\cmark & \cmark & \cmark & \cmark & \textbf{83.4} & \textbf{80.3} & \textbf{74.8} & \textbf{85.9} \\
\bottomrule[1.5pt]
\end{tabular}
\end{adjustbox}

\vspace{-2pt}
\end{wraptable}

\noindent\textbf{Which caption facets support action prediction?}
To quantify the contribution of different caption components, we ablate over caption combinations in Table \ref{tab:action-ablation-or-caption-types}. Adding global captions improves necessity and category AUROC on MOVER, although the gains are less consistent on VitalDB. Incorporating action captions yields the strongest overall results, improving necessity and category AUROC on both datasets relative to supervision without the action facet. These results indicate that action-relevant language provides useful auxiliary supervision for learning sensor-action associations.

\begin{wraptable}{r}{0.55\linewidth}
\vspace{-9pt}
\centering
\renewcommand{\arraystretch}{1.1}
\setlength{\tabcolsep}{8pt}

\caption{\small{\textbf{Ablation of caption generation and action conditioning.}
AUROC$^\uparrow$ (\%) is used as the metric.}}
\label{tab:action-ablation-or-decoder}

\vspace{-6pt}
\small
\begin{adjustbox}{max width=\linewidth}
\begin{tabular}{l cc cc}
\toprule[1.5pt]
\multirow{2}{*}{\textbf{Method}}
& \multicolumn{2}{c}{\textbf{MOVER}}
& \multicolumn{2}{c}{\textbf{VitalDB}} \\
\cmidrule(lr){2-3}\cmidrule(lr){4-5}
& Necessity & Category & Necessity & Category \\
\midrule
Action only & 78.1 & 72.1 & 54.1 & 72.8 \\
No cond. & \underline{80.4} & \underline{77.2} & \underline{66.3} & \underline{84.4} \\
No decoder & 78.3 & 76.4 & 65.1 & 84.1 \\
\midrule
\rowcolor{gray!15}
\oursbase & \textbf{83.4} & \textbf{80.3} & \textbf{74.8} & \textbf{85.9} \\
\bottomrule[1.5pt]
\end{tabular}
\end{adjustbox}

\vspace{-8pt}
\end{wraptable}

\noindent\textbf{Is caption supervision alone sufficient?}
Finally, we separate the benefit of language supervision from the mechanism used to connect language with action prediction. The action-only variant removes caption generation and its associated loss; the no-action-conditioning variant retains caption generation but removes conditioning on predicted actions; and the no-decoder variant retains language supervision but generates captions directly from the LLM backbone. Table \ref{tab:action-ablation-or-decoder} verifies that the full model achieves the highest AUROC across all four endpoints. Both caption-generating variants outperform action-only training, confirming the benefit of language supervision, while explicit action conditioning and the multimodal fusion decoder provide additional improvements.

\section{Discussion}
\label{sec:conclusion}

\textbf{Limitation.}
\ours has not yet undergone clinical validation and is not intended for direct clinical decision-making. Real-world deployment would require prospective validation, safety evaluation, and appropriate regulatory review. Our current implementation trains separate models for each domain, in part because their action spaces are defined independently. An important next step is to develop a generalist SLA model that operates across domains through shared representations of heterogeneous actions. In addition, our experiments assume a fixed set of sensor modalities and input channels, whereas real-world sensing systems vary across devices, signal availability, and acquisition settings. Extending SLA to robustly handle such sensor heterogeneity remains an important direction for future work.

\textbf{Conclusion.} We present Sensor-Language-Action (SLA) modeling, extending sensor intelligence from understanding what is observed toward reasoning about the actions those observations support. SLA uses language as a semantic interface that connects sensor observations, language, and actions within a shared modeling framework. We instantiate this paradigm via a large-scale benchmark, multi-faceted language supervision, and \ours, achieving strong performance across action prediction, state understanding, and action explanation. Further analyses show generalization to unseen actions and datasets, semantic and continuous structure in the learned action space, and predictive information about future sensor states.  
Together, SLA provides a foundation for general sensor models that jointly reason over sensing, language, and action.

\section*{Acknowledgments}
We gratefully acknowledge the support of NIH R21-EB038753, the Amazon Research Awards, the NVIDIA Academic Grant, and the Biswas Foundation. Any opinions, findings, conclusions, or recommendations expressed in this material are those of the authors and do not necessarily reflect the views of the funders.

\bibliography{ref}
\bibliographystyle{plainnat}

\newpage
\appendix
\section{Details of Datasets}
\label{sec:appendix:dataset}
\subsection{Dataset Splits and Statistics}
\label{sec:appendix:dataset-splits}
\label{sec:appendix:captioning:statistics}

\textbf{Dataset usage.} Our development datasets include MC-MED~\citep{kansal2025mcmed} and MIMIC-III~\citep{johnson2016mimic} for clinical care, MOVER~\citep{samad2023mover} and VitalDB~\citep{lee2022vitaldb} for operating room (OR) care, and MetaboNet~\citep{wolff2026metabonet} and PEDAP~\citep{wadwa2023pedap} for continuous glucose monitoring (CGM). All datasets above contribute to training and evaluation splits. We additionally use MIMIC-IV~\citep{PhysioNet-mimiciv-2.0} dataset as an external transfer cohort, which is fully unseen during training. Table~\ref{tab:cohort-composition} summarizes the cohort sizes, with a combined size of 116K across the seven datasets. Table~\ref{tab:dataset-splits} reports the resulting training, validation, and test pools after quality filtering for action evaluation. Across the six development datasets, the training pool contains more than 2.3 million windows. For evaluation, test windows are balanced with respect to action necessity; For test evaluation, we balance-sample windows with and without a target action, and apply the same strategy to 1,000 MIMIC-IV windows for external evaluation.

\begin{table*}[h]
\renewcommand{\arraystretch}{1.1}
\setlength{\tabcolsep}{6pt}
\centering
\caption{\textbf{Dataset usage and split sizes.}}
\vspace{-5pt}
\label{tab:dataset-splits}
\small

\begin{adjustbox}{width=\textwidth}
\begin{tabular}{@{}l rr rr rr r@{}}
\toprule[1.5pt]
\multirow{2}{*}{\textbf{Split}} & \multicolumn{2}{c}{\textbf{Clinical}}
& \multicolumn{2}{c}{\textbf{Operating Room}}
& \multicolumn{2}{c}{\textbf{CGM}}
& \multicolumn{1}{c}{\textbf{External}} \\
\cmidrule(lr){2-3}
\cmidrule(lr){4-5}
\cmidrule(lr){6-7}
\cmidrule(lr){8-8}

{}
& \multicolumn{1}{c}{MC-MED~\cite{kansal2025mcmed}}
& \multicolumn{1}{c}{MIMIC-III~\cite{johnson2016mimic}}
& \multicolumn{1}{c}{MOVER~\cite{samad2023mover}}
& \multicolumn{1}{c}{VitalDB~\cite{lee2022vitaldb}}
& \multicolumn{1}{c}{MetaboNet~\cite{wolff2026metabonet}}
& \multicolumn{1}{c}{PEDAP~\cite{wadwa2023pedap}}
& \multicolumn{1}{c}{MIMIC-IV~\cite{PhysioNet-mimiciv-2.0}} \\
\midrule
\midrule
Train      & 90,412 & 16,184 & 113,865 & 48,800 & 1,945,570 & 127,437 & -- \\
Validation &    403 &    129 &     238 &  1,000 &    10,000 &  10,000 & -- \\
Test       &  2,753 &  3,402 &     795 &  2,000 &    10,000 &  10,000 & 1,000 \\
\bottomrule[1.5pt]
\end{tabular}
\end{adjustbox}

\end{table*}

\begin{table}[!ht]
\renewcommand{\arraystretch}{1.1}
\setlength{\tabcolsep}{6pt}
\centering
\caption{\textbf{Full cohort sizes across the seven datasets.}}
\vspace{-5pt}
\label{tab:cohort-composition}
\small
\begin{adjustbox}{width=\textwidth}
\begin{tabular}{@{}l rrr rr rr r@{}}
\toprule[1.5pt]
\multirow{2}{*}{\textbf{Dataset}} & \multicolumn{3}{c}{\textbf{Clinical}} & \multicolumn{2}{c}{\textbf{Operating Room}} & \multicolumn{2}{c}{\textbf{CGM}} & \multirow{2}{*}{\textbf{Total}} \\
\cmidrule(lr){2-4}\cmidrule(lr){5-6}\cmidrule(lr){7-8}
{}
& MC-MED~\cite{kansal2025mcmed} & MIMIC-III~\cite{johnson2016mimic} & MIMIC-IV~\cite{PhysioNet-mimiciv-2.0}
& MOVER~\cite{samad2023mover} & VitalDB~\cite{lee2022vitaldb}
& MetaboNet~\cite{wolff2026metabonet} & PEDAP~\cite{wadwa2023pedap}
& \\
\midrule
\midrule
\textbf{Cohort size} & 49,196 & 35,276 & 122 & 26,329 & 3,657 & 1,095 & 65 & \textbf{115,740} \\
\bottomrule[1.5pt]
\end{tabular}
\end{adjustbox}
\end{table}

\textbf{Sensor Modalities and Coverage.}
\label{sec:appendix:sensor-modalities}
We use waveform signals and numerical measurements as sensor inputs across domains. As summarized in Table~\ref{tab:sensor-modalities}, waveform modalities include ECG, plethysmography, respiration, invasive pressure, capnography, EEG, and CGM glucose traces, while numerical inputs include vital signs, hemodynamic and ventilator measurements, laboratory values, and historical insulin delivery. We further perform dataset-specific quality control and sensor harmonization to ensure reliable model inputs. In particular, we have reconciled cross-dataset modality aliases and acquisition variants and filtered low-quality waveform windows where applicable. The resulting sample-level coverage is reported in Table~\ref{tab:sensor-modalities}. Waveform and numerical inputs provide broad and complementary availability across datasets, with near-complete coverage for several cohorts and sensor types.

\begin{table*}[t]
\renewcommand{\arraystretch}{1.1}
\setlength{\tabcolsep}{6pt}
\centering
\caption{\textbf{Sensor modalities and training-set coverage for development datasets.}}
\vspace{-5pt}
\label{tab:sensor-modalities}
\small

\begin{adjustbox}{width=\textwidth}
\begin{tabularx}{\textwidth}{@{}l>{\raggedright\arraybackslash}Xcc@{}}
\toprule[1.5pt]
\textbf{Sensor Type} & \textbf{Modalities} & \multicolumn{2}{c}{\textbf{Training Coverage (\%)}} \\
\midrule
\midrule
\multicolumn{2}{@{}l}{\textit{\textbf{Clinical}}} & MC-MED~\cite{kansal2025mcmed} & MIMIC-III~\cite{johnson2016mimic} \\
\midrule
Waveform & ECG leads; plethysmography; respiration; arterial and other invasive-pressure waveforms. & 70.8 & 78.8 \\
\addlinespace[3pt]
Numerical & Vital signs (HR, RR, SpO$_2$, blood pressure, temperature); hemodynamics; ventilator measurements; oxygen delivery; glucose; neurological and pain scores; urine output. & 99.0 & 42.9 \\
\midrule
\multicolumn{2}{@{}l}{\textit{\textbf{Operating Room (OR)}}} & MOVER~\cite{samad2023mover} & VitalDB~\cite{lee2022vitaldb} \\
\midrule
Waveform & ECG leads; plethysmography; arterial and central venous pressure; capnography; airway pressure; EEG. & 25.1 & 100.0 \\
\addlinespace[3pt]
Numerical & Vital signs; invasive and non-invasive blood pressure; cardiac output and related hemodynamics; ventilation and respiratory gases; BIS and EEG-derived indices; laboratory measurements; urine output. & 99.6 & 100.0 \\
\midrule
\multicolumn{2}{@{}l}{\textit{\textbf{Continuous Glucose Monitoring (CGM)}}} & MetaboNet~\cite{wolff2026metabonet} & PEDAP~\cite{wadwa2023pedap} \\
\midrule
Waveform & Glucose time series. & 100.0 & 100.0 \\
\addlinespace[3pt]
Numerical & Historical basal-insulin delivery. & 100.0 & 100.0 \\
\bottomrule[1.5pt]
\end{tabularx}
\end{adjustbox}

\end{table*}

\subsection{Sensor-Action Sample Construction}
\label{sec:appendix:captioning:samples}

\noindent\textbf{Input State.} Each sample is anchored at a decision time $T_0$ and combines a historical sensor window with textual individual context. The input windows span 30 minutes for clinical care, 5 minutes for operating room (OR) care, and 2 hours for CGM. The CGM model input contains 24 five-minute slots preceding the decision slot. Text context records available individual background, presentation, and prior actions, depending on the source dataset. The recorded target action is stored separately as supervision.

\noindent\textbf{Action Targets.} The action hierarchy addresses whether to act (necessity), which type of action to select (category), and which particular action to take (fine-grained label), where applicable. We derive supervision for these levels from recorded actions within the target window. Clinical and operating room (OR) samples can have multiple category and label targets. The CGM task predicts bolus occurrence and its dose bin, without a separate fine-grained label vocabulary. Each sample pairs a historical sensor window with action targets defined relative to the decision time $T_0$. Table~\ref{tab:action-target-construction} summarizes the positive and negative sample construction rules and the action types across domains.

\begin{table}[!ht]
\renewcommand{\arraystretch}{1.1}
\setlength{\tabcolsep}{6pt}
\centering
\caption{\textbf{Action targets and negative-sample construction.}}
\vspace{-5pt}
\label{tab:action-target-construction}
\small
\begin{adjustbox}{width=\textwidth}
\begin{tabularx}{\textwidth}{@{}>{\raggedright\arraybackslash}p{0.14\textwidth}>{\raggedright\arraybackslash}p{0.21\textwidth}>{\raggedright\arraybackslash}p{0.25\textwidth}>{\raggedright\arraybackslash}X@{}}
\toprule[1.5pt]
\textbf{Dataset} & \textbf{Target interval} & \textbf{Positive definition} & \textbf{Negative construction} \\
\midrule
\midrule
MC-MED,\newline MIMIC-III & [$T_0$, $T_0$ + 5 min] & At least one mapped clinical action. & No mapped action; retained and history-matched evaluation controls. \\
\midrule
MOVER & ($T_0$, $T_0$ + 2 min] & A mapped medication, procedure, device, or diagnostic event. & No mapped event; same-grid, within-case sampling preferentially matched by procedure progress. \\
\addlinespace[4pt]
VitalDB & ($T_0$, $T_0$ + 2 min] & An explicit command or control change. & No mapped change; uncertain initial or post-gap observations are excluded. \\
\midrule
MetaboNet,\newline PEDAP & Decision slot $T_0$ & A canonical insulin-bolus onset. & No onset; within-individual sampling excludes recent boluses and overlapping histories. \\
\bottomrule[1.5pt]
\end{tabularx}
\end{adjustbox}
\end{table}

\noindent\textbf{Action Category.} Source-specific action names and codes are mapped to categories and fine-grained labels based on clinical practice. Clinical and operating room (OR) samples may contain multiple actions, with repeated occurrences of the same label counted once within the target window. For CGM, necessity denotes bolus occurrence, and positive boluses are assigned to four mutually exclusive dose categories: (0, 1), [1, 2), [2, 4), and [4, $\infty$) U. The full action space contain 39 Clinical categories, 17 operating room (OR) categories, and four CGM dose categories, totaling \numaction domain-specific action categories. Table~\ref{tab:action-vocabulary-support} summarizes the number of action targets and their positive test-window support across domains, reported as minimum, median, and maximum counts.

\begin{table}[H]
\renewcommand{\arraystretch}{1.1}
\setlength{\tabcolsep}{6pt}
\centering
\caption{\textbf{Number of action categories and positive-sample support.} Support is reported as the minimum / median / maximum number of positive samples across targets.}
\vspace{-5pt}
\label{tab:action-vocabulary-support}
\small

\begin{adjustbox}{width=\textwidth}
\begin{tabular*}{\textwidth}{@{\extracolsep{\fill}}lcccc@{}}
\toprule[1.5pt]
\multirow{2}{*}{\textbf{Dataset}} & \multicolumn{2}{c}{\textbf{Action Category}}
& \multicolumn{2}{c}{\textbf{Positive Window Support}} \\
\cmidrule(lr){2-3}\cmidrule(lr){4-5}

{}
& \multicolumn{1}{c}{\textbf{Categories}}
& \multicolumn{1}{c}{\textbf{Fine-grained labels}}
& \multicolumn{1}{c}{\textbf{Category}}
& \multicolumn{1}{c}{\textbf{Fine-grained label}} \\
\midrule
\midrule
MC-MED~\cite{kansal2025mcmed}
& 29 & 56 & 12 / 41 / 132 & 10 / 18.5 / 132 \\

MIMIC-III~\cite{johnson2016mimic}
& 28 & 60 & 12 / 75.5 / 437 & 10 / 43 / 263 \\

\midrule
MOVER~\cite{samad2023mover}
& 13 & 22 & 10 / 21 / 48 & 10 / 13 / 28 \\

VitalDB~\cite{lee2022vitaldb}
& 5 & 14 & 23 / 219 / 831 & 23 / 218.5 / 343 \\

\midrule
MetaboNet~\cite{wolff2026metabonet}
& 4 & N/A & 1,166 / 1,274 / 1,382 & N/A \\

PEDAP~\cite{wadwa2023pedap}
& 4 & N/A & 273 / 886.5 / 3,180 & N/A \\

\bottomrule[1.5pt]
\end{tabular*}
\end{adjustbox}

\end{table}

\section{Multi-Faceted Caption Construction}
\label{sec:appendix:captioning}

We construct caption targets through structured fact extraction followed by rule-based selection and text rendering. For each decision time $T_0$, the pipeline extracts findings from the preceding sensor window and available individual context. Intermediate records retaining for each finding its measurement identity, numerical summary, temporal information, and source. These records are organized into four complementary caption facets.

\subsection{Observational Caption Construction}
\label{sec:appendix:captioning:aspects}
\label{sec:appendix:captioning:observational}

Observational captions describe individual measurements and waveform channels. For numerical observations, we summarized statistics such as the median, range, and number of observations, and associated these summaries with measurement-specific state descriptors. Sparsely sampled measurements are reported by their observed values. For waveforms, channel-specific extractors produce rate, periodicity, amplitude, and pattern descriptors where supported. waveform-derived estimates are reported separately from charted measurements.

The selected findings are rendered with their measurement names and numerical summaries. When signal quality is insufficient, the caption notes limited availability instead of reporting the corresponding descriptors. For CGM, the summary covers the glucose level, range, trend, and latest reading, with range-occupancy and variability descriptions added when applicable.

\subsection{Relational Caption Construction}
\label{sec:appendix:captioning:relational}

Relational captions describe temporal structure and relationships between physiological measurements. Temporal descriptions are derived from the extracted trajectories, distinguishing directional changes from persistent or relatively unchanged states and are produced only when the observations are sufficient to support a trend. Where historical event records are available, the captions also summarize their timing and duration, such as earlier carbohydrate entries, insulin delivery, or basal-rate changes in CGM.

Cross-channel descriptions use predefined, physiologically related channel pairs rather than unrestricted pairwise correlations. For example numerical vital signs and their corresponding waveforms. A joint statement is generated only when the relevant observations are available and satisfy the corresponding compatibility and quality rules.

\subsection{Global Caption Construction}
\label{sec:appendix:captioning:global}

Global captions combine available individual context with selected physiological findings. Context includes recorded background information, presentation, and relevant history. The summary is chosen from a predefined set of state descriptions according to the findings that support each candidate.

For clinical and operating room (OR) data, candidate states are ranked by the support provided by available findings and are presented together with their main supporting evidence. CGM uses predefined combinations of glucose dynamics and historical insulin or carbohydrate information to identify state descriptions. When no state description is sufficiently supported, the caption explicitly reports this.

\subsection{Action Caption Construction}
\label{sec:appendix:captioning:action}

Action captions connect sensor observations and individual context with action-relevant evidence. We use predefined matching rules to select relevant findings from the extracted facts and summarize how they support or qualify an action-related interpretation. Where applicable, recorded action categories guide this selection, while the underlying evidence is drawn from pre-decision observations. These descriptions are intended as evidence associations, not as causal explanations or treatment recommendations.

Finally, the selected facts are verbalized using sentence templates and phrase banks that vary wording while retaining measurement identities, values, and evidence qualifications. Example captions are provided in Appendix~\ref{sec:appendix:multi-faceted-caption-examples}, and the corresponding template pools are listed in Appendix~\ref{sec:appendix:template-pool}.

\section{Method Implementation Details}
\label{sec:appendix:method}

\subsection{\ours Implementation}
\label{sec:appendix:method:architecture}

\noindent\textbf{Sensor Encoding.} We process waveform and numerical inputs with separate sensor encoders. For the Clinical and OR domains, waveform signals are encoded using a pretrained DINO encoder adapted from~\citep{shuai2026osf}, which remains frozen during SLA training. For CGM, we don't use a large sensor encoder because the signal length is smaller. Instead, we use a trainable convolutional patchifier to convert glucose traces into temporal tokens. Numerical measurements are encoded with a trainable Transformer, where each observation is represented by combining its original embedding with measurement-identity and temporal embeddings.

\noindent\textbf{Structured Action Heads.} All three heads operate on the backbone hidden state at the final input position, denoted as \(\mathbf h \in \mathbb R^d\), where \(d\) is the language hidden dimension. \ding{182} \textit{Necessity heads.}
Independent binary linear heads predict the presence of each action-necessity type and are trained with binary cross-entropy $\mathcal L_{\mathrm{nec}}$.
\ding{183} \textit{Category heads.}
Independent binary linear heads predict the action categories. They are trained with positive-class-weighted binary cross-entropy $\mathcal L_{\mathrm{cat}}$.
\ding{184} \textit{Fine-grained label heads.}
Within each action category, we obtain text embeddings for the candidate fine-grained actions and project them together with $\mathbf h$ into a shared embedding space. Candidate actions are scored by cosine similarity and trained with cross-entropy within each category, denoted as $\mathcal L_{\mathrm{label}}$.

Together, the training objective $\mathcal L_{\mathrm{act}}$ used in Section~\ref{sec:method} can be written as:
\begin{equation}
\mathcal L_{\mathrm{act}}
=\lambda_{\mathrm{nec}}\mathcal L_{\mathrm{nec}}
+\lambda_{\mathrm{cat}}\mathcal L_{\mathrm{cat}}
+\lambda_{\mathrm{label}}\mathcal L_{\mathrm{label}},
\label{eq:appendix:action-loss}
\end{equation}
where $\lambda_{\mathrm{nec}}$ weights the action-necessity loss, $\lambda_{\mathrm{cat}}$ is for category prediction, and $\lambda_{\mathrm{label}}$ is the weight of fine-grained label loss.

\noindent\textbf{Hierarchical Memory.} The Hierarchical Memory pathway uses two query stages. Local queries aggregate encoder features within each channel-time group, while temporal queries aggregate the resulting local memories across time within the same channel. Within each branch, local query parameters are shared across channel-time groups, and temporal query parameters are shared across channels. Channel embeddings are added at both query stages, with time embeddings additionally added to the local queries. The local, temporal, and global memory tokens are concatenated as the final memory representation. Each query stage consists of multi-head cross-attention followed by a residual feed-forward block. Waveform and numerical inputs use separate parameters, feature widths, and query budgets. For numerical inputs, observations are grouped by measurement identity and relative timestamp, and the per-measurement summary tokens are fed into the query hierarchy and are appended directly to the numerical memory.

\noindent\textbf{Multi-modal Fusion Decoder.} Action tokens $\mathbf P$ represent the predicted action profile, including an action-necessity token and action-category tokens. Each token combines the corresponding action embedding with a projected prediction score. In the hierarchical configuration, waveform and numerical selectors retrieve relevant sensor tokens using the output hidden state $\mathbf H$ from the LLM backbone, together with the pooled action profile $\mathbf P$. Waveform candidates are obtained from patch-level sensor features, while numerical candidates are obtained from observation tokens. Selected candidates are projected into the language hidden space and weighted by their retrieval scores to form sensor evidence $\mathbf S$. The action-profile branch $f_p$ and sensor-evidence branch $f_s$ each contain cross-attention, a feed-forward block, and an output projection. Both operate on the original caption states $\mathbf H$, attending to the action profile $\mathbf P$ and sensor evidence $\mathbf S$. The refined caption states $\mathbf H'$ are then passed through the backbone's original language-model head. Training uses ground-truth caption prefixes and predicted actions with weighted autoregressive cross-entropy loss, where signal-related caption tokens have been assigned a higher loss weight.

\subsection{Baseline and Ablation Details}
\label{sec:appendix:method:baselines}

\noindent\textbf{Supervised Baselines.} We compare ConvTrans~\citep{foumani2024improving} and MMIM~\citep{han2021mmim} with \ours in the main action tables. We adapt MMIM to combine sensor recordings with language context using a trainable CNN and frozen BERT encoder~\citep{devlin2019bert}. Their representations are fused through an MLP and passed to task-specific action classifiers. We retain the mutual-information and contrastive objectives from the original implementation, and jointly optimize them with supervised action prediction, and do not use a caption-generation objective.

\textbf{TS Foundation Models.} We linearly probe Chronos-2~\citep{ansari2025chronos2}, Moirai~\citep{woo2024moirai}, and GluFormer~\citep{lutsker2026foundation} for action prediction. Sensor sequences are encoded into pretrained representations and pooled before being passed to linear prediction heads, while the pretrained encoders remain frozen. These sensor-only baselines do not receive language context.

\textbf{Zero-shot multi-modal LLMs.}
We perform zero-shot inference with state-of-the-art multi-modal LLMs, including GPT-5.6-Luna~\citep{openai2026gpt56luna} and DeepSeek-V3.2~\citep{deepseek2025v32}, through a CodeAct-style interface~\citep{wang2024codeact}. Instead of serializing long sensor recordings into the prompt, the models execute Python code to inspect pre-decision measurements and compute relevant summaries. Language context and tool outputs are then used for action prediction and evidence-caption generation. Action prompts provide the available action categories and the candidate fine-grained labels within each category. We directly evaluate the returned discrete predictions without model adaptation or probability-based scoring.

\noindent\textbf{TS Language Models.} \ding{182} \textbf{OpenTSLM.} To accommodate the full sensor histories used by \ours, we replace OpenTSLM's native sensor encoder with our domain-specific sensor encoders. We retain the original SP and Flamingo fusion module and initialize their remaining components from their official LLama-based checkpoints~\citep{langer2026opentslm}. Specifically, we use the model variant pretrained on ECG signals for clinical and operating room (OR) tasks, and use the one pretrained on general time-series for the CGM domain. \ding{183} \textbf{SensorLM.} We adapt SensorLM~\citep{zhang2025sensorlm} to our datasets with minimal changes to its original formulation. It uses the same sensor encoders as \ours, together with a unimodal text encoder and a multimodal decoder that cross-attends to sensor features.

\noindent\textbf{Ablation Variants.} We construct ablations based on the configuration of \oursbase. \ding{182} \textbf{Caption supervision and conditioning.} The action-only variant removes caption generation and its training loss while retaining the action predictors. The no-decoder variant retains caption supervision but generates text directly from the language backbone without the residual fusion decoder. The no-action-conditioning variant retains the decoder but removes its predicted-action inputs. \ding{183} \textbf{Language-backbone adaptation.} We compare two alternatives to LoRA-based adaptation. The LLaVA-style variant~\citep{liu2023visual} freezes the language backbone and maps sensor features into input tokens through trainable projectors, while the Flamingo-style variant~\citep{alayrac2022flamingo} injects sensor features through gated cross-attention modules in the frozen backbone. Both retain the same sensor encoders, action heads, and caption supervision, with all task-specific modules trainable except the pretrained DINO encoder.

\subsection{Training Details}
\label{sec:appendix:method:training}

Table~\ref{tab:training-settings} summarizes optimization settings and effective batch sizes including gradient accumulation. 
\begin{table*}[t]
\renewcommand{\arraystretch}{1.1}
\setlength{\tabcolsep}{6pt}
\centering
\caption{\textbf{Training configurations for OpenSLA and adapted sensor-language baselines.}}
\vspace{-5pt}
\label{tab:training-settings}
\small

\begin{adjustbox}{width=\textwidth}
\begin{tabularx}{\textwidth}{>{\raggedright\arraybackslash}p{0.37\textwidth}*{3}{>{\centering\arraybackslash}X}}
\toprule[1.5pt]
\textbf{Setting} & \textbf{\ours} & \textbf{OpenTSLM}~\cite{langer2026opentslm} & \textbf{SensorLM}~\cite{zhang2025sensorlm} \\
\midrule
\midrule
\multicolumn{4}{@{}l}{\textit{\textbf{Optimization:}}} \\
Optimizer & AdamW & AdamW & AdamW \\
Peak learning rate & $5\times10^{-5}$ & $5\times10^{-5}$ & $5\times10^{-5}$ \\
Weight decay & 0.01 & 0.01 & 0.01 \\
Warmup fraction & 0.10 & 0.10 & 0.10 \\
Learning-rate decay & Linear & Linear & Linear \\
Gradient clipping norm & 0.5 & 0.5 & 0.5 \\
LoRA rank / $\alpha$ / dropout & 16 / 32 / 0.05 & 16 / 32 / 0.05 & -- \\
\midrule
\multicolumn{4}{@{}l}{\textit{\textbf{Effective Batch Size:}}} \\
Clinical & 16 & 16 & 64 \\
Operating room (OR) & 16 & 16 & 64 \\
CGM & 16 & 16 & 256 \\
\bottomrule[1.5pt]
\end{tabularx}
\end{adjustbox}

\end{table*}

\section{Task Setup}
\label{sec:appendix:task-setup}

\subsection{Action Prediction}
\label{sec:appendix:task-setup:action}

We evaluate action prediction at three levels corresponding to whether to act, which action category to select, and which fine-grained action to take. Predictions are compared with held-out action annotations derived from the source records. Clinical and operating room (OR) tasks support all three levels. CGM necessity indicates a bolus intervention, and its categories correspond to dose bins; there is no separate fine-grained label task. Text-generating baselines are evaluated by mapping their generated actions to the same target vocabulary.

We report AUROC from continuous prediction scores and balanced accuracy (BAcc), defined as the mean of sensitivity and specificity. For the main action tables, thresholds maximize validation BAcc separately for each binary target, retaining the calibration scope of each evaluated model. Fine-grained labels are evaluated within ground-truth-positive category cohorts, averaging eligible labels within each category and then across categories.

The $\pm$ values summarize bootstrap uncertainty as half the width of 95\% percentile intervals from 2,000 individual-level cluster resamples (surgical cases for VitalDB), displayed symmetrically around the point estimates. Checkpoints and evaluation thresholds remain fixed; these values reflect evaluation-cohort sampling uncertainty rather than variation across training runs.

\subsection{Caption-Based Evaluation}
\label{sec:appendix:task-setup:caption}

\noindent\textbf{State Understanding.} We extract physiological quantities from generated captions and compare them with reference statistics computed from the observed history. Table~\ref{tab:caption-evaluation-protocol} specifies the quantity, statistic, and readout for each task. We compute mean absolute error (MAE) on reference-eligible samples with a valid extracted prediction, using the reference unit.

\noindent\textbf{Action Explanation.} We compare generated text with the action-evidence portion of the reference caption using BERTScore F1. We use \texttt{distilbert-base-uncased}. The reported result is the mean per-sample F1, multiplied by 100.

\begin{table*}[t]
\renewcommand{\arraystretch}{1.1}
\setlength{\tabcolsep}{6pt}
\centering
\caption{\textbf{Caption evaluation tasks and scoring rules.} Definitions for the sensor-state and action-evidence tasks reported in the main results.}
\vspace{-5pt}
\label{tab:caption-evaluation-protocol}
\small
\begin{adjustbox}{width=\textwidth}
\begin{tabularx}{\textwidth}{@{}>{\raggedright\arraybackslash}p{0.18\textwidth}>{\raggedright\arraybackslash}p{0.24\textwidth}>{\raggedright\arraybackslash}X>{\centering\arraybackslash}p{0.14\textwidth}@{}}
\toprule[1.5pt]
\textbf{Quantity / Task} & \textbf{Reference Statistic} & \textbf{Prediction Readout} & \textbf{Metric} \\
\midrule
\midrule
\multicolumn{4}{@{}l}{\textit{\textbf{Clinical: 30-minute sensor history}}} \\
\midrule
Cardiac index & Window median\newline (L/min/m$^2$) & Extract an explicit cardiac-index median. & MAE$\downarrow$ \\
\addlinespace[3pt]
ECG-II rate & ECG-derived rate proxy\newline (beats/min) & Extract the rate attributed to ECG lead II. & MAE$\downarrow$ \\
\addlinespace[3pt]
RESP rate & Respiration-waveform rate proxy\newline (breaths/min) & Extract a rate attributed to the RESP waveform. & MAE$\downarrow$ \\
\midrule
\multicolumn{4}{@{}l}{\textit{\textbf{Operating room: 5-minute sensor history;}}} \\
\midrule
Respiratory rate & Window median\newline (breaths/min) & Extract an explicit median of measured RR. & MAE$\downarrow$ \\
\addlinespace[3pt]
Temperature & Window median\newline ($^\circ$C) & Extract an explicit temperature median. & MAE$\downarrow$ \\
\addlinespace[3pt]
End-tidal CO$_2$ & Window median\newline (mmHg) & Extract an explicit ETCO$_2$ median. & MAE$\downarrow$ \\
\midrule
\multicolumn{4}{@{}l}{\textit{\textbf{CGM: 2-hour pre-decision history}}} \\
\midrule
Basal rate & Mean recorded basal-insulin rate\newline (U/h) & Extract a mean/average basal rate. & MAE$\downarrow$ \\
\addlinespace[3pt]
Carbohydrate amount & Amount of the latest recorded carbohydrate event\newline (g) & Extract an amount paired with a pre-decision recency in the preceding 120 minutes, and choose the latest unambiguous event. & MAE$\downarrow$ \\
\midrule
\multicolumn{4}{@{}l}{\textit{\textbf{Action-evidence text: all six development datasets}}} \\
\midrule
Action-evidence agreement & Ground-truth action-evidence text & Compare generated captions, against the same reference evidence. & BERTScore\newline F1$\uparrow$~\cite{zhang2019bertscore} \\
\bottomrule[1.5pt]
\end{tabularx}
\end{adjustbox}
\end{table*}

\subsection{Unseen Action Classification}
\label{sec:appendix:task-setup:unseen}

To examine generalization to unseen fine-grained actions within a known category, we construct a balanced binary classification set from MIMIC-III. The target is Imipenem/Cilastatin, a single antibiotic label absent from the supervised action training targets, while the antibiotic category is seen during training.

\noindent\textbf{Zero-Shot Action Scoring.} \ours represents candidate actions through textual descriptors rather than label-specific classifier weights. At inference, we encode the unseen descriptor, \texttt{antibiotic: Imipenem/Cilastatin}, by averaging its language-backbone token embeddings and applying the trained descriptor projection. Its cosine similarity with the projected multi-modal decision representation provides the fine-grained action score. All model parameters remain fixed; no additional training on the unseen label is required.

\subsection{External-Cohort Evaluation}
\label{sec:appendix:task-setup:external}

We evaluate the Clinical models on waveform-linked MIMIC-IV samples without further adaptation to this cohort. Each sample provides pre-decision sensor history and language context, with actions mapped to the Clinical target vocabulary. We report necessity and category AUROC and BAcc; discrete action predictions from text-generating baselines contribute BAcc but not AUROC. This experiment measures transfer to another MIMIC release and cohort.

The evaluation cohort contains 1,000 waveform-qualified windows from 87 individuals, with 500 action-positive and 500 action-negative samples. Both classes use a common five-minute candidate grid and a five-minute target interval. Sampling approximately matches action prevalence to the MIMIC-III reference cohort while balancing pre-decision availability characteristics.

\begingroup
\raggedbottom
\setlength{\intextsep}{5pt plus 2pt minus 2pt}
\section{Additional Results}
\label{sec:appendix:additional-results}

We provide additional results and analysis including sensor-language adaptation, caption supervision, generalization, and the information retained in learned representations.

\subsection{Sensor--Language Adaptation}

We compare direct LoRA adaptation with two alternatives: a LLaVA-style variant that freezes the language backbone and learns sensor projectors, and a Flamingo-style variant that injects trainable cross-attention into the frozen backbone. Both retain the same action heads and caption supervision. As shown in Table~\ref{tab:action-ablation-or-fusion}, direct LoRA achieves the highest necessity and category AUROC on both MOVER and VitalDB datasets.

\begin{table}[H]
\renewcommand{\arraystretch}{1.1}
\setlength{\tabcolsep}{6pt}
\centering
\caption{\textbf{Sensor--language adaptation.} Necessity and category AUROC (\%) on operating room datasets.}
\vspace{-5pt}
\label{tab:action-ablation-or-fusion}
\small
\begin{adjustbox}{width=\textwidth}
\begin{tabularx}{\textwidth}{>{\raggedright\arraybackslash}p{0.32\textwidth}*{4}{>{\centering\arraybackslash}X}}
\toprule[1.5pt]
\multirow{3.5}{*}{\textbf{Method}} & \multicolumn{2}{c}{\textbf{MOVER}} & \multicolumn{2}{c}{\textbf{VitalDB}} \\
\cmidrule(lr){2-3}\cmidrule(lr){4-5}
 & Necessity & Category & Necessity & Category \\
\cmidrule(lr){2-2}\cmidrule(lr){3-3}\cmidrule(lr){4-4}\cmidrule(lr){5-5}
 & AUROC$^\uparrow$ & AUROC$^\uparrow$ & AUROC$^\uparrow$ & AUROC$^\uparrow$ \\
\midrule
\midrule
{LLaVA-style}~\cite{liu2023visual} & 51.5 & 66.4 & 54.4 & 60.0 \\
{Flamingo-style}~\cite{alayrac2022flamingo} & \underline{82.9} & \underline{76.1} & \underline{68.9} & \underline{82.1} \\
\midrule
\rowcolor{gray!15}
\ours & \textbf{83.4} & \textbf{80.3} & \textbf{74.8} & \textbf{85.9} \\
\bottomrule[1.5pt]
\end{tabularx}
\end{adjustbox}
\end{table}

\subsection{Unseen Action Generalization}

We further evaluate zero-shot prediction of Imipenem in the clinical domain, a fine-grained antibiotic label excluded from supervised action training. \ours scores this unseen label using its textual representation without additional training. As shown in Table~\ref{tab:unseen-imipenem}, \ourshq achieves the highest F1 scores and balanced accuracies.

\begin{table}[H]
\renewcommand{\arraystretch}{1.1}
\setlength{\tabcolsep}{6pt}
\centering
\caption{\textbf{Unseen-action prediction results.}}
\vspace{-5pt}
\label{tab:unseen-imipenem}
\small
\begin{adjustbox}{width=\textwidth}
\begin{tabularx}{\textwidth}{>{\raggedright\arraybackslash}p{0.45\textwidth}*{2}{>{\centering\arraybackslash}X}}
\toprule[1.5pt]
\multirow{2}{*}{\textbf{Method}} & \multicolumn{2}{c}{\textbf{Imipenem}} \\
\cmidrule(lr){2-3}
& \multicolumn{1}{c}{F1$^\uparrow$} & \multicolumn{1}{c}{BAcc$^\uparrow$} \\
\midrule
\midrule
{OpenTSLM}~\cite{langer2026opentslm} & 17.4 & 54.8 \\
\midrule
{GPT-5.6-Luna}~\cite{openai2026gpt56luna} & 42.9 & \underline{61.9} \\
{DeepSeek-V3.2}~\cite{deepseek2025v32} & 34.5 & 54.8 \\
\midrule
\rowcolor{gray!15}
\ours & \textbf{70.0} & \textbf{71.4} \\
\bottomrule[1.5pt]
\end{tabularx}
\end{adjustbox}
\end{table}

\subsection{Future-State Readout}

We test whether current-state representations retain information about subsequent glucose dynamics. On the MetaboNet dataset, we construct paired current and future two-hour signals and extract the glucose mean, minimum, maximum, and slope over the following two hours. We then fit ridge regression on the frozen representation from each model and evaluate it with five-fold cross-validation. It is fitted using nested cross-validation, with five outer folds and three inner folds for regularization selection. Chronos-2 is evaluated separately using its native zero-shot forecasts.

Table~\ref{tab:cgm-future-state} reports glucose-summary MAE in mg/dL and slope MAE in mg/dL/min. The \ours variants have lower error on mean, minimum, and maximum glucose than the listed baselines. Slope errors are close: \ours, OpenTSLM-SP, and SensorLM all round to 0.39. These results concern a trained readout of frozen representations, not zero-shot generation of future glucose values.

\begin{table}[H]
\renewcommand{\arraystretch}{1.1}
\setlength{\tabcolsep}{6pt}
\centering
\caption{\textbf{Future-state readout on MetaboNet.} MAE for the next two hours: mean, minimum, maximum, and slope.}
\vspace{-5pt}
\label{tab:cgm-future-state}
\small
\begin{adjustbox}{width=\textwidth}
\begin{tabularx}{\textwidth}{>{\raggedright\arraybackslash}p{0.32\textwidth}*{4}{>{\centering\arraybackslash}X}}
\toprule[1.5pt]
\multirow{2}{*}{\textbf{Method}} & \multicolumn{4}{c}{\textbf{MetaboNet}} \\
\cmidrule(lr){2-5}
{}
& Mean$^\downarrow$
& Min$^\downarrow$
& Max$^\downarrow$
& Slope$^\downarrow$ \\
\midrule
\midrule
{OpenTSLM}~\cite{langer2026opentslm}
& 26.5 & 24.2 & 30.6 & 0.39 \\
{SensorLM}~\cite{zhang2025sensorlm}
& 26.9 & 24.6 & 30.5 & 0.39 \\
{Chronos-2}~\cite{ansari2025chronos2}
& 29.7 & 29.3 & 32.8 & 0.48 \\
{GluFormer}~\cite{lutsker2026foundation}
& 33.2 & 30.0 & 38.7 & 0.41 \\
\midrule
\rowcolor{gray!15}
\oursbase
& \underline{24.0}
& \textbf{22.3}
& \underline{28.0}
& \underline{0.39} \\
\rowcolor{gray!15}
\ourshq
& \textbf{24.0}
& \underline{22.5}
& \textbf{27.7}
& \textbf{0.39} \\
\bottomrule[1.5pt]
\end{tabularx}
\end{adjustbox}
\end{table}

\subsection{Contribution of Sensor Inputs}
\label{sec:appendix:sensor-inputs}

We compare \ours with a context-only variant that removes waveform and numerical inputs while retaining the same pre-decision text context, hierarchical action heads, supervision text captions, and action-conditioned fusion decoder. As shown in Table~\ref{tab:appendix:context-only}, including sensor inputs improves necessity and category AUROC on both datasets.

\begin{table}[H]
\renewcommand{\arraystretch}{1.1}
\setlength{\tabcolsep}{6pt}
\centering
\caption{\textbf{Contribution of sensor inputs.} We report the results (AUROC (\%)) of action necessity and category prediction.}
\vspace{-5pt}
\label{tab:appendix:context-only}
\small
\begin{adjustbox}{width=\textwidth}
\begin{tabularx}{\textwidth}{>{\raggedright\arraybackslash}p{0.32\textwidth}*{4}{>{\centering\arraybackslash}X}}
\toprule[1.5pt]
\multirow{3.5}{*}{\textbf{Method}} & \multicolumn{2}{c}{\textbf{MOVER}} & \multicolumn{2}{c}{\textbf{VitalDB}} \\
\cmidrule(lr){2-3}\cmidrule(lr){4-5}
 & Necessity & Category & Necessity & Category \\
\cmidrule(lr){2-2}\cmidrule(lr){3-3}\cmidrule(lr){4-4}\cmidrule(lr){5-5}
 & AUROC$^\uparrow$ & AUROC$^\uparrow$ & AUROC$^\uparrow$ & AUROC$^\uparrow$ \\
\midrule
\midrule
Context-only & \underline{79.8} & \underline{75.7} & \underline{64.9} & \underline{83.4} \\
\midrule
\rowcolor{gray!15}
\ours & \textbf{83.4} & \textbf{80.3} & \textbf{74.8} & \textbf{85.9} \\
\bottomrule[1.5pt]
\end{tabularx}
\end{adjustbox}
\end{table}

\begin{figure}[!htbp]
\centering
\includegraphics[width=0.94\linewidth]{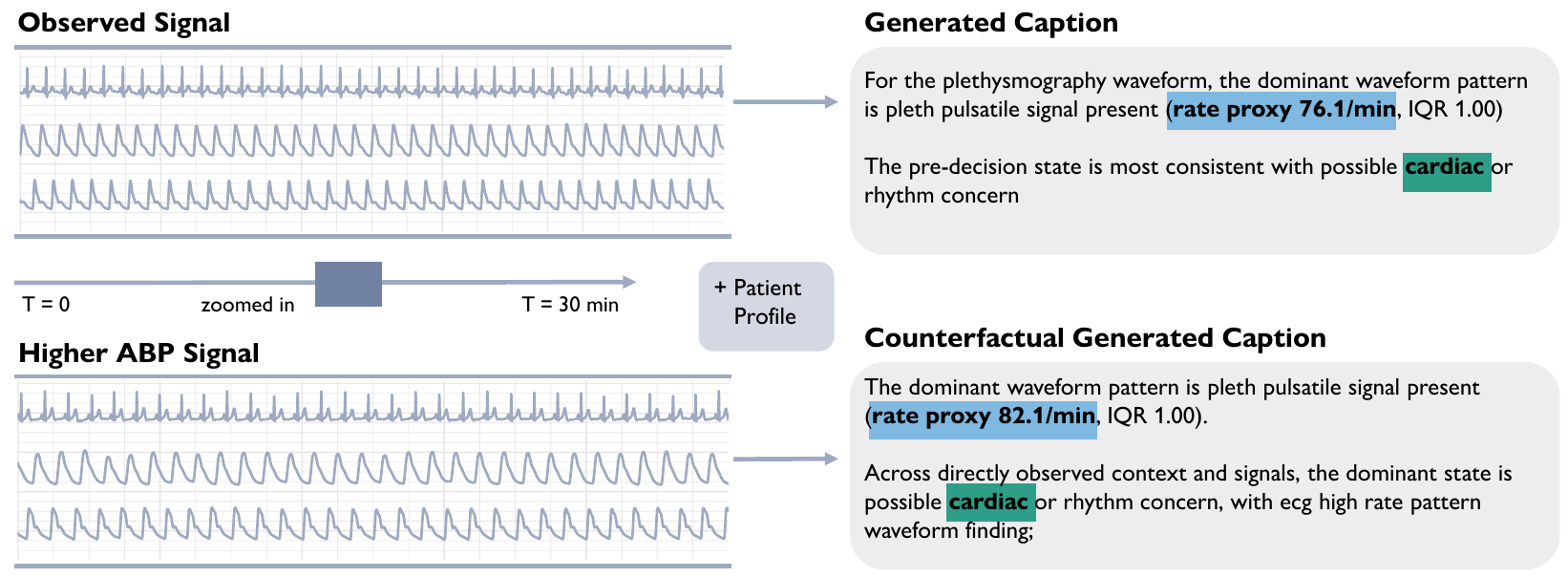}
\caption{\textbf{Caption responses to sensor substitution.} Patient context is held fixed while the supplied sensor recording changes.}
\label{fig:counterfactual}
\end{figure}

\subsection{Controlled Sensor Substitution}

We hold the patient context fixed and replace the observed sensor recording with a higher-ABP example. As shown in Fig.~\ref{fig:counterfactual}, this sensor substitution changes the generated caption: the Pleth rate proxy increases from 76.1 to 82.1/min, and the state summary additionally identifies a high-rate ECG pattern, while the overall cardiac/rhythm concern remains unchanged. This controlled comparison demonstrates that the generated caption responds to changes in the sensor input, while remaining the patient context part unchanged.

\subsection{Hierarchical Memory Efficiency}

Table~\ref{tab:hcsm-efficiency} compares \oursbase and \ourshq on paired validation windows from each domain, with the values to the right of each arrow corresponding to \ourshq. We use batch size 1 on a GH200 GPU and measure forward latency for the sensor encoders, backbone, and necessity/category heads on the action prediction task; compression and speedup are computed as ratios of domain-level means. Hierarchical Memory reduces both token counts and forward latency on average in the Clinical and OR domains. For CGM, sensor sequences are already short, so the additional memory-processing overhead largely offsets the savings from compression, resulting in little runtime benefit.

\begin{table}[H]
\renewcommand{\arraystretch}{1.1}
\setlength{\tabcolsep}{6pt}
\centering
\caption{\textbf{Hierarchical Memory efficiency.}}
\vspace{-5pt}
\label{tab:hcsm-efficiency}
\small
\begin{adjustbox}{width=\textwidth}
\begin{tabularx}{\textwidth}{>{\raggedright\arraybackslash}p{0.18\textwidth}*{3}{>{\centering\arraybackslash}X}}
\toprule[1.5pt]
\textbf{Domain} & Sensor Tokens$^\downarrow$ & Compression$^\uparrow$ & Speedup$^\uparrow$ \\
\midrule
\midrule
Clinical & $5801.0 \rightarrow 400.4$ & $14.49\times$ & $3.4\times$ \\
OR & $3915.8 \rightarrow 838.0$ & $4.67\times$ & $1.7\times$ \\
CGM & $33.0 \rightarrow 21.0$ & $1.57\times$ & $1\times$ \\
\bottomrule[1.5pt]
\end{tabularx}
\end{adjustbox}
\end{table}

\endgroup

\begingroup
\raggedbottom
\section{Additional Examples}
\label{sec:appendix:additional-examples}
\subsection{Multi-faceted Caption Examples}
\label{sec:appendix:multi-faceted-caption-examples}
Examples 3 and 4 are re-rendered with consistent waveform-evidence filtering; the remaining examples reproduce saved reference captions.
\newcommand{\exampleheading}[1]{\par\addvspace{10pt}\noindent\makebox[\linewidth]{\leaders\hbox{=}\hfill\enspace Example #1\enspace\leaders\hbox{=}\hfill}\par\nobreak\vspace{5pt}}
\newcommand{\captionpart}[2]{\par\addvspace{6pt}\noindent\textbf{#1:} #2\par}
\newcommand{\captionfield}[1]{\textbf{#1:}}
\begin{tcolorbox}[
  enhanced,
  breakable,
  sharp corners,
  colback=black!4,
  colframe=black,
  colbacktitle=black,
  coltitle=white,
  boxrule=0.4pt,
  title={Examples Captions Generated by Our Data Pipeline},
  center title,
  fonttitle=\normalfont\small,
  fontupper=\fontsize{9.5}{11.5}\selectfont\ttfamily,
  left=7pt,right=7pt,top=3pt,bottom=7pt,
  boxsep=0pt,
  toptitle=5pt,bottomtitle=5pt,
  before skip=7pt,after skip=7pt,
  parbox=false
]
\raggedright
\exampleheading{1}
\captionpart{OBSERVATIONAL}{\captionfield{Numeric} 1-minute heart-rate variability was evidence of hrv abnormal (1min\_HRV median 10, range 5-32, n=19) and persistently outside the expected range. The heart rate series indicates tachycardia and is persistently outside the expected range (HR median 110, range 107-120, n=19). \captionfield{Waveform} The Pleth trace did not meet the quality threshold for morphology or rate-proxy statements. Because Resp produced incompatible rhythm or periodicity proxies, it was excluded from the waveform conclusion. Lead II waveform evidence includes tachycardic ECG waveform proxy (rate proxy 108.8/min, IQR 0.20).}
\captionpart{RELATIONAL}{\captionfield{Temporal} Temporal pattern: short-window HRV is abnormal with little recent change across the current decision window; evidence of hrv abnormal. Near the decision point, heart-rate channel shows elevated heart rate and is abnormal but not changing much. \captionfield{Cross-signal} elevated cardiac rate across numeric and ECG evidence is present across the corresponding signal channels.}
\captionpart{GLOBAL}{\captionfield{Context} Before the decision point, the contextual record contains reported shortness of breath at presentation. \captionfield{State summary} Across directly observed context and signals, the dominant state is possible cardiac or rhythm concern finding, with hrv abnormal context; rapid cardiac rate; ecg high rate pattern waveform finding.}
\captionpart{ACTION}{elevated heart-rate pattern (HR median 110, range 107-120, n=19); trend: abnormal but not changing much; elevated cardiac rate across numeric and ECG evidence; cardiac waveform channel showed high-rate Lead II periodicity (rate proxy 108.8/min, IQR 0.20) is relevant to circulatory support, although the available evidence is indirect or incomplete.}

\exampleheading{2}
\captionpart{OBSERVATIONAL}{\captionfield{Numeric} Across the current observation interval, perfusion signal shows poor perfusion signal; Perf median 1, range 1-1, n=30; persistently outside the expected range. For respiratory rate, the dominant finding is elevated respiratory rate; RR median 24, range 17-29, n=30. \captionfield{Waveform} The Pleth trace did not meet the quality threshold for morphology or rate-proxy statements. The rate or periodicity proxies from Resp were internally inconsistent, so that channel was not used for waveform interpretation. ECG channel summary: ecg rate proxy normal; rate proxy 81.2/min, IQR 0.10.}
\captionpart{RELATIONAL}{\captionfield{Temporal} Temporal pattern: Perf is sustained without a clear directional shift across the half-hour signal window; poor perfusion signal. Over the monitored interval, RR is abnormal but not changing much with elevated respiratory rate.}
\captionpart{GLOBAL}{\captionfield{Context} Available contextual evidence includes reported neurologic at presentation. \captionfield{State summary} The pre-decision state is most consistent with possible hemodynamic instability signal; key evidence is reduced peripheral perfusion proxy; weak Pleth waveform.}
\captionpart{ACTION}{A possible reason for the infectious diagnostic testing workup is the pre-T0 finding tachypnea (RR median 24, range 17-29, n=30); trend: persistently outside the expected range; perfusion concern (Perf median 1, range 1-1, n=30); trend: persistently outside the expected range; peripheral pulse waveform showed weak pulse waveform amplitude (IQR 0.00); no direct note statement confirms it.}

\exampleheading{3}
\captionpart{OBSERVATIONAL}{\captionfield{Numeric} The heart rate series indicates tachycardia and is abnormal with little recent change (HR median 107, range 102-111, n=17). \captionfield{Waveform} Because Pleth was only partly usable in this window, no morphology or rate inference was drawn from it. The available Resp trace was present, but its proxy features disagreed, so no single morphology or rate interpretation was reported. ECG Lead II: nonuniform cardiac waveform proxy; rate proxy 99.9/min, IQR 0.35.}
\captionpart{RELATIONAL}{\captionfield{Temporal} Over the monitored interval, heart-rate channel is persistently outside the expected range with elevated heart-rate pattern.}
\captionpart{GLOBAL}{\captionfield{Context} Presentation context includes chief complaint of hypertension or hypotension. \captionfield{State summary} The pre-decision state is most consistent with possible cardiac or rhythm concern context; key evidence is elevated heart-rate pattern; irregular cardiac waveform pattern.}
\captionpart{ACTION}{The available state shows elevated heart rate (HR median 107, range 102-111, n=17); trend: outside the expected range without a strong trend; because this recorded treatment can represent context-specific care, these facts are described as background for cardiorenal volume management rather than a definitive rationale.}

\exampleheading{4}
\captionpart{OBSERVATIONAL}{\captionfield{Numeric} For heart-rate channel, the dominant finding is elevated heart-rate pattern; HR median 100, range 95-108, n=30. \captionfield{Waveform} Proxy interpretation was deferred for Pleth because the available waveform was quality-constrained. Waveform availability for Resp was adequate, yet the derived proxies were discordant and therefore not interpreted as a single finding. The cardiac waveform channel shows variable ECG waveform periodicity; rate proxy 95.2/min, IQR 0.40.}
\captionpart{RELATIONAL}{\captionfield{Temporal} Temporal pattern: heart rate is abnormal but not changing much across the current observation interval; heart rate above expected range.}
\captionpart{GLOBAL}{\captionfield{Context} Pre-decision context includes arrival complaint includes psych or substance; reported medication or refill at presentation. \captionfield{State summary} The pre-decision state is most consistent with possible cardiac or rhythm concern signal; key evidence is tachycardia; irregular Lead II structure.}
\captionpart{ACTION}{Because no direct note reason was available, heart rate above expected range (HR median 100, range 95-108, n=30); trend: sustained without a clear directional shift is retained as a possible indication for evaluation of altered mental status or neurologic concern.}

\exampleheading{5}
\captionpart{OBSERVATIONAL}{\captionfield{Numeric} The heart rate has limited charting that supports elevated heart-rate pattern (HR median 101, range 101-101, n=1); this is based on one pre-decision HR reading, without a reliable trajectory. rapid breathing pattern appears in one pre-decision RR reading (RR median 23, range 23-23, n=1); no temporal trend is assigned. \captionfield{Waveform} ECG channel summary: ECG high-rate proxy; rate proxy 100.0/min, IQR 1.00. For Plethysmography waveform, the dominant waveform pattern is uneven Pleth pulsatility (rate proxy 91.3/min, IQR 1.00).}
\captionpart{GLOBAL}{\captionfield{Context} Before the decision point, the contextual record contains presentation mentions bleeding or gi bleed. \captionfield{State summary} Across directly observed context and signals, the dominant state is possible infection or sepsis concern signal, with rapid cardiac rate; RR above expected range.}
\captionpart{ACTION}{A secondary signal for symptom-control intervention is elevated heart-rate pattern (HR median 101, range 101-101, n=1; one pre-decision HR reading); no stronger domain-specific fact was selected.}

\exampleheading{6}
\captionpart{OBSERVATIONAL}{\captionfield{Numeric} Basal delivery was zero throughout the prior two hours. \captionfield{Waveform} Glucose during the preceding two hours stayed within 71-100 mg/dL, median 88 mg/dL, inside the 70-180 mg/dL range. In the two-hour window, the fitted slope of sensor glucose was +6 mg/dL per hour; glucose was flat, within 10 mg/dL per hour. The last reading before the decision was 94 mg/dL, 5 minutes before it.}
\captionpart{RELATIONAL}{\captionfield{Temporal} Over the two-hour window before the decision, no bolus, carbohydrate entry, small delivery run or basal change was logged.}
\captionpart{GLOBAL}{\captionfield{Context} Patient context: 48 yr old, a male patient, 104 kg body weight, height of 170 cm, body mass index 36.0 kg/m2, duration of diabetes 38 years, automated insulin delivery. \captionfield{State summary} Pre-decision state: readings in 70-180 mg/dL throughout, changing under 10 mg/dL per hour.}
\captionpart{ACTION}{Findings on the side of no bolus: pump basal of zero entering the decision.}

\exampleheading{7}
\captionpart{OBSERVATIONAL}{\captionfield{Numeric} No basal delivery was recorded in the prior two hours. \captionfield{Waveform} Glucose over the two-hour observation window had a median of 97 mg/dL, in the target range of 70-180 mg/dL, and went from a low of 78 to a high of 113 mg/dL. Glucose over the two-hour window before the decision: decreasing, -18 mg/dL per hour. Glucose at the last reading, 5 minutes before the decision, was 78 mg/dL.}
\captionpart{RELATIONAL}{\captionfield{Temporal} In the two-hour lookback: no bolus, carbohydrate entry, small delivery run or basal change.}
\captionpart{GLOBAL}{\captionfield{Context} Patient profile: 28 yr old, a male patient, a weight of 58 kg, 175 cm tall, BMI equal to 18.8 kg/m2, 11 years since diagnosis, automated insulin delivery (AID). \captionfield{State summary} Summary of the state before the decision: none of the defined states applied; evidence was limited.}
\captionpart{ACTION}{Against a bolus: no pump basal in the slot before the decision.}

\exampleheading{8}
\captionpart{OBSERVATIONAL}{\captionfield{Numeric} Basal insulin stayed at zero for all of the last two hours. Insulin on board, slot before the decision: 0.13 U. \captionfield{Waveform} Sensor glucose in the preceding two hours: median 114 mg/dL, in the 70-180 mg/dL target range, lowest 101 and highest 122 mg/dL. Glucose over the two hours before the decision: stable (under 10 mg/dL per hour either way), -8 mg/dL per hour. The last CGM value was 101 mg/dL, 5 minutes before the decision.}
\captionpart{RELATIONAL}{\captionfield{Temporal} No timed events (bolus, carbohydrate, small delivery run, basal change) in the last two hours.}
\captionpart{GLOBAL}{\captionfield{Context} Known patient attributes: aged 43, male, a closed-loop system. \captionfield{State summary} State before the decision: a 70-180 mg/dL trace with a slope below 10 mg/dL per hour.}
\captionpart{ACTION}{Findings opposing an insulin bolus: zero basal on the pump just before the decision.}

\exampleheading{9}
\captionpart{OBSERVATIONAL}{\captionfield{Numeric} Basal insulin stayed at zero for all of the two-hour window. \captionfield{Waveform} CGM readings in the last two hours spanned 88-136 mg/dL, with a median of 117 mg/dL, neither below 70 nor above 180 mg/dL. Over the two-hour period before the decision, glucose was dropping, at -20 mg/dL per hour. The reading closest to the decision was 88 mg/dL, 5 minutes earlier.}
\captionpart{RELATIONAL}{\captionfield{Temporal} The record lists no bolus, carbohydrate entry, small delivery run or basal change for the two-hour period before the decision.}
\captionpart{GLOBAL}{\captionfield{Context} Patient record: 35 years old, of female sex, body weight 54 kg, 157 cm tall, BMI at 21.9 kg/m2, living with diabetes 30 years, automated insulin delivery. \captionfield{State summary} Overall state before the decision: no selected glycaemic state; direct evidence was limited.}
\captionpart{ACTION}{Pre-decision evidence against an insulin bolus: pump basal at zero just before the decision.}

\exampleheading{10}
\captionpart{OBSERVATIONAL}{\captionfield{Numeric} A constant basal rate of 0.90 U/h over the two hours up to the decision delivered 1.80 U. Insulin on board immediately before the decision was 0.01 U. \captionfield{Waveform} Median sensor glucose over the two-hour window was 150 mg/dL, in the target range of 70-180 mg/dL; readings 142-153 mg/dL. For the two-hour period before the decision, glucose had a fitted slope of -5 mg/dL per hour and was flat (less than 10 mg/dL per hour). Latest glucose: 142 mg/dL at 5 minutes before the decision.}
\captionpart{RELATIONAL}{\captionfield{Temporal} No timed events (bolus, carbohydrate, small delivery run, basal change) in the two hours prior to the decision.}
\captionpart{GLOBAL}{\captionfield{Context} Patient details on record: aged 66, recorded as male, weighing 108 kg, a height of 185 cm, computed BMI 31.4 kg/m2, 26 years of diabetes, an insulin pump with CGM. \captionfield{State summary} Overall glycaemic state: glucose kept inside 70-180 mg/dL, slope under 10 mg/dL per hour.}
\captionpart{ACTION}{No finding for or against an insulin bolus was selected from the observed window.}
\end{tcolorbox}
\subsection{Template pool}
\label{sec:appendix:template-pool}
\begin{tcolorbox}[
  enhanced,
  breakable,
  sharp corners,
  colback=black!4,
  colframe=black,
  colbacktitle=black,
  coltitle=white,
  boxrule=0.4pt,
  title={OBSERVATIONAL},
  center title,
  fonttitle=\normalfont\small,
  fontupper=\fontsize{8.2}{10.2}\selectfont\ttfamily,
  left=7pt,right=7pt,top=5pt,bottom=7pt,
  boxsep=0pt,
  toptitle=5pt,bottomtitle=5pt,
  before skip=7pt,after skip=7pt,
  parbox=false
]
\raggedright\sloppy\setlength{\parindent}{0pt}\setlength{\parskip}{0pt}
\noindent NUMERIC\_CAPTION\_TEMPLATES = [\par
\noindent \hspace*{1em}"\{modality\_phrase\} was \{numeric\_finding\} (\{stat\_phrase\}) and \{trend\_phrase\}.",\par
\noindent \hspace*{1em}"Across \{window\_phrase\}, \{modality\_phrase\} shows \{numeric\_finding\}; \{stat\_phrase\}; \{trend\_phrase\}.",\par
\noindent \hspace*{1em}"The \{modality\_phrase\} series indicates \{numeric\_finding\} and is \{trend\_phrase\} (\{stat\_phrase\}).",\par
\noindent \hspace*{1em}"Recorded \{modality\_phrase\} values support \{numeric\_finding\}; \{stat\_phrase\}; \{trend\_phrase\}.",\par
\noindent \hspace*{1em}"For \{modality\_phrase\}, the dominant finding is \{numeric\_finding\}; \{stat\_phrase\}.",\par
\noindent \hspace*{1em}"\{numeric\_finding\} is present in the \{modality\_phrase\} series and is \{trend\_phrase\}.",\par
\noindent ]\par
\smallskip
\noindent NUMERIC\_OBSERVATION\_ONLY\_TEMPLATES = [\par
\noindent \hspace*{1em}"Observed \{modality\_phrase\} during \{window\_phrase\} (\{stat\_phrase\}).",\par
\noindent \hspace*{1em}"The current \{modality\_phrase\} summary is \{stat\_phrase\}.",\par
\noindent \hspace*{1em}"Recorded \{modality\_phrase\} values in \{window\_phrase\} are summarized as \{stat\_phrase\}.",\par
\noindent ]\par
\smallskip
\noindent NUMERIC\_SPARSE\_TEMPLATES = [\par
\noindent \hspace*{1em}"\{modality\_phrase\} has limited charting that supports \{numeric\_finding\} (\{stat\_phrase\}); this is based on \{sparse\_observation\_phrase\}, without a reliable trajectory.",\par
\noindent \hspace*{1em}"\{numeric\_finding\} appears in \{sparse\_observation\_phrase\} (\{stat\_phrase\}); no temporal trend is assigned.",\par
\noindent \hspace*{1em}"Evidence from \{sparse\_observation\_phrase\} is consistent with \{numeric\_finding\} (\{stat\_phrase\}); the values are retained but temporal detail is withheld because observations are sparse.",\par
\noindent \hspace*{1em}"The available values for \{modality\_phrase\} indicate \{numeric\_finding\} (\{stat\_phrase\}), although the evidence consists of only \{sparse\_observation\_phrase\}.",\par
\noindent ]\par
\smallskip
\noindent LAB\_RESULT\_CAPTION\_TEMPLATES = [\par
\noindent \hspace*{1em}"\{modality\_phrase\}: \{numeric\_finding\}; \{stat\_phrase\}.",\par
\noindent \hspace*{1em}"\{numeric\_finding\} appears in recent \{modality\_phrase\} results (\{stat\_phrase\}).",\par
\noindent \hspace*{1em}"Recent lab evidence from \{modality\_phrase\} shows \{numeric\_finding\}; \{stat\_phrase\}.",\par
\noindent \hspace*{1em}"The resulted \{modality\_phrase\} panel contributes \{numeric\_finding\}: \{stat\_phrase\}.",\par
\noindent \hspace*{1em}"\{stat\_phrase\} supports \{numeric\_finding\} in the \{modality\_phrase\} category.",\par
\noindent \hspace*{1em}"Lab context before the decision point includes \{numeric\_finding\} from \{modality\_phrase\}; \{stat\_phrase\}.",\par
\noindent \hspace*{1em}"\{modality\_phrase\} result summary: \{numeric\_finding\}; \{stat\_phrase\}.",\par
\noindent \hspace*{1em}"A recent \{modality\_phrase\} finding is \{numeric\_finding\}, recorded as \{stat\_phrase\}.",\par
\noindent ]\par
\smallskip
\noindent WAVEFORM\_GENERIC\_TEMPLATES = [\par
\noindent \hspace*{1em}"\{modality\_phrase\}: \{pattern\_phrase\}; \{stat\_phrase\}.",\par
\noindent \hspace*{1em}"The \{modality\_phrase\} signal shows \{pattern\_phrase\} (\{stat\_phrase\}).",\par
\noindent \hspace*{1em}"\{modality\_phrase\} provides waveform evidence of \{pattern\_phrase\}; \{stat\_phrase\}.",\par
\noindent \hspace*{1em}"For \{modality\_phrase\}, the dominant waveform pattern is \{pattern\_phrase\} (\{stat\_phrase\}).",\par
\noindent ]\par
\smallskip
\noindent WAVEFORM\_ECG\_II\_TEMPLATES = [\par
\noindent \hspace*{1em}"ECG Lead II: \{pattern\_phrase\}; \{stat\_phrase\}.",\par
\noindent \hspace*{1em}"Lead II waveform evidence includes \{pattern\_phrase\} (\{stat\_phrase\}).",\par
\noindent \hspace*{1em}"The cardiac waveform channel shows \{pattern\_phrase\}; \{stat\_phrase\}.",\par
\noindent \hspace*{1em}"ECG channel summary: \{pattern\_phrase\}; \{stat\_phrase\}.",\par
\noindent ]\par
\smallskip
\noindent WAVEFORM\_PLETH\_TEMPLATES = [\par
\noindent \hspace*{1em}"Pleth waveform: \{pattern\_phrase\}; \{stat\_phrase\}.",\par
\noindent \hspace*{1em}"The pulse-ox waveform shows \{pattern\_phrase\} (\{stat\_phrase\}).",\par
\noindent \hspace*{1em}"Peripheral pulse waveform summary: \{pattern\_phrase\}; \{stat\_phrase\}.",\par
\noindent \hspace*{1em}"Pleth channel evidence is summarized as \{pattern\_phrase\}, with \{stat\_phrase\}.",\par
\noindent ]\par
\smallskip
\noindent WAVEFORM\_RESP\_TEMPLATES = [\par
\noindent \hspace*{1em}"Resp waveform: \{pattern\_phrase\}; \{stat\_phrase\}.",\par
\noindent \hspace*{1em}"The respiration waveform shows \{pattern\_phrase\} (\{stat\_phrase\}).",\par
\noindent \hspace*{1em}"Breathing waveform evidence includes \{pattern\_phrase\}; \{stat\_phrase\}.",\par
\noindent \hspace*{1em}"Respiration waveform summary: \{pattern\_phrase\}; \{stat\_phrase\}.",\par
\noindent ]\par
\smallskip
\noindent CGM\_WAVEFORM\_CAPTION\_TEMPLATES = \{\par
\noindent \hspace*{1em}"glucose\_summary": [\par
\noindent \hspace*{2em}"Median sensor glucose over \{period\} was \{median\} mg/dL, \{band\_phrase\}; readings \{lo\}-\{hi\} mg/dL.",\par
\noindent \hspace*{2em}"Over \{period\}, median glucose was \{median\} mg/dL, \{band\_phrase\}, with readings from \{lo\} to \{hi\} mg/dL.",\par
\noindent \hspace*{1em}],\par
\noindent \hspace*{1em}"glucose\_trend": [\par
\noindent \hspace*{2em}"Glucose was \{direction\} over \{period\}, with a fitted slope of \{slope\} mg/dL per hour.",\par
\noindent \hspace*{2em}"The fitted glucose slope over \{period\} was \{slope\} mg/dL per hour; glucose was \{direction\}.",\par
\noindent \hspace*{1em}],\par
\noindent \hspace*{1em}"glucose\_last": [\par
\noindent \hspace*{2em}"The last reading before the decision was \{last\} mg/dL, \{m\} minutes before it.",\par
\noindent \hspace*{2em}"The most recent glucose, \{m\} minutes before the decision, was \{last\} mg/dL.",\par
\noindent \hspace*{1em}],\par
\noindent \hspace*{1em}"time\_in\_ranges": [\par
\noindent \hspace*{2em}"Time in ranges over \{period\}: \{p\_lt54\}\% below 54, \{p\_lt70\}\% below 70, \{p\_70\_180\}\% at 70-180, \{p\_gt180\}\% above 180 and \{p\_gt250\}\% above 250 mg/dL.",\par
\noindent \hspace*{2em}"Of valid readings in \{period\}, \{p\_70\_180\}\% were at 70-180 mg/dL, \{p\_lt70\}\% below 70 (\{p\_lt54\}\% below 54) and \{p\_gt180\}\% above 180 (\{p\_gt250\}\% above 250).",\par
\noindent \hspace*{1em}],\par
\noindent \hspace*{1em}"variability": [\par
\noindent \hspace*{2em}"The p10-p90 glucose spread over \{period\} was \{spread\} mg/dL, at least 60 mg/dL.",\par
\noindent \hspace*{2em}"Glucose variability in \{period\}: p10-p90 spread \{spread\} mg/dL, 60 mg/dL or more.",\par
\noindent \hspace*{1em}],\par
\noindent \hspace*{1em}"no\_glucose": [\par
\noindent \hspace*{2em}"No usable sensor glucose was available in \{period\}.",\par
\noindent \hspace*{2em}"Sensor glucose was unavailable for \{period\}.",\par
\noindent \hspace*{1em}],\par
\noindent \}\par
\smallskip
\noindent CGM\_NUMERIC\_CAPTION\_TEMPLATES = \{\par
\noindent \hspace*{1em}"basal\_zero\_throughout": [\par
\noindent \hspace*{2em}"No basal insulin was delivered in \{period\}.",\par
\noindent \hspace*{2em}"Basal delivery was zero throughout \{period\}.",\par
\noindent \hspace*{1em}],\par
\noindent \hspace*{1em}"basal\_steady": [\par
\noindent \hspace*{2em}"Basal insulin ran at a constant \{rate\} U/h over \{period\}, \{total\} U in total.",\par
\noindent \hspace*{2em}"Over \{period\}, basal delivery held at \{rate\} U/h, totalling \{total\} U.",\par
\noindent \hspace*{1em}],\par
\noindent \hspace*{1em}"basal\_varying": [\par
\noindent \hspace*{2em}"Basal insulin over \{period\} averaged \{mean\} U/h, ranging \{min\}-\{max\} U/h, \{total\} U in total.",\par
\noindent \hspace*{2em}"Over \{period\}, basal delivery ranged \{min\}-\{max\} U/h (mean \{mean\} U/h), totalling \{total\} U.",\par
\noindent \hspace*{1em}],\par
\noindent \hspace*{1em}"iob": [\par
\noindent \hspace*{2em}"Insulin on board before the decision was \{iob\} U.",\par
\noindent \hspace*{2em}"Just before the decision, insulin on board was \{iob\} U.",\par
\noindent \hspace*{1em}],\par
\noindent \hspace*{1em}"cob": [\par
\noindent \hspace*{2em}"Carbohydrate on board before the decision was \{cob\} g.",\par
\noindent \hspace*{2em}"Just before the decision, carbohydrate on board was \{cob\} g.",\par
\noindent \hspace*{1em}],\par
\noindent \}\par
\smallskip
\end{tcolorbox}
\begin{tcolorbox}[
  enhanced,
  breakable,
  sharp corners,
  colback=black!4,
  colframe=black,
  colbacktitle=black,
  coltitle=white,
  boxrule=0.4pt,
  title={RELATIONAL},
  center title,
  fonttitle=\normalfont\small,
  fontupper=\fontsize{8.2}{10.2}\selectfont\ttfamily,
  left=7pt,right=7pt,top=5pt,bottom=7pt,
  boxsep=0pt,
  toptitle=5pt,bottomtitle=5pt,
  before skip=7pt,after skip=7pt,
  parbox=false
]
\raggedright\sloppy\setlength{\parindent}{0pt}\setlength{\parskip}{0pt}
\noindent TEMPORAL\_CAPTION\_TEMPLATES = [\par
\noindent \hspace*{1em}"Temporal pattern: \{modality\_phrase\} is \{trend\_phrase\} across \{window\_phrase\}; \{numeric\_finding\}.",\par
\noindent \hspace*{1em}"Over the monitored interval, \{modality\_phrase\} is \{trend\_phrase\} with \{numeric\_finding\}.",\par
\noindent \hspace*{1em}"The observed trajectory for \{modality\_phrase\} is \{trend\_phrase\}; \{stat\_phrase\}.",\par
\noindent \hspace*{1em}"Near the decision point, \{modality\_phrase\} shows \{numeric\_finding\} and is \{trend\_phrase\}.",\par
\noindent ]\par
\smallskip
\noindent CROSS\_SIGNAL\_CAPTION\_TEMPLATES = [\par
\noindent \hspace*{1em}"\{numeric\_finding\} and \{waveform\_finding\} are directionally consistent.",\par
\noindent \hspace*{1em}"\{numeric\_finding\} is \{connector\_evidence\} \{waveform\_finding\}.",\par
\noindent \hspace*{1em}"\{waveform\_finding\} provides waveform support for \{numeric\_finding\}.",\par
\noindent \hspace*{1em}"\{numeric\_finding\} is not clearly matched by waveform evidence because \{uncertainty\_phrase\}.",\par
\noindent \hspace*{1em}"\{cross\_signal\_finding\}: \{numeric\_finding\} \{connector\_additive\} \{waveform\_finding\}.",\par
\noindent \hspace*{1em}"Together, \{numeric\_finding\} and \{waveform\_finding\} support \{clinical\_state\}.",\par
\noindent \hspace*{1em}"\{cross\_signal\_finding\} is present, based only on available numeric and waveform evidence.",\par
\noindent \hspace*{1em}"\{waveform\_finding\} and \{numeric\_finding\} point to \{clinical\_state\}, with proxy-level certainty.",\par
\noindent \hspace*{1em}"\{numeric\_finding\} contrasts with \{waveform\_finding\}; signal interpretation is \{uncertainty\_phrase\}.",\par
\noindent \hspace*{1em}"Cross-signal evidence: \{cross\_signal\_finding\}; \{supporting\_findings\}.",\par
\noindent ]\par
\smallskip
\noindent CROSS\_SIGNAL\_EXPLICIT\_FACT\_TEMPLATES = [\par
\noindent \hspace*{1em}"\{cross\_signal\_finding\} is supported by the paired numeric and waveform observations.",\par
\noindent \hspace*{1em}"Matched numeric and waveform channels show \{cross\_signal\_finding\}.",\par
\noindent \hspace*{1em}"\{cross\_signal\_finding\} is present across the corresponding signal channels.",\par
\noindent ]\par
\smallskip
\noindent CGM\_TEMPORAL\_CAPTION\_TEMPLATES = \{\par
\noindent \hspace*{1em}"bolus\_event": [\par
\noindent \hspace*{2em}"A \{dose\} U bolus was recorded \{m\} minutes before the decision.",\par
\noindent \hspace*{2em}"Bolus: \{dose\} U, \{m\} minutes before the decision.",\par
\noindent \hspace*{1em}],\par
\noindent \hspace*{1em}"carb\_event": [\par
\noindent \hspace*{2em}"\{grams\} g of carbohydrate was recorded \{m\} minutes before the decision.",\par
\noindent \hspace*{2em}"Carbohydrate: \{grams\} g, \{m\} minutes before the decision.",\par
\noindent \hspace*{1em}],\par
\noindent \hspace*{1em}"small\_deliveries\_one": [\par
\noindent \hspace*{2em}"One run of deliveries under 0.25 U per row (\{total\} U) was not counted as a bolus.",\par
\noindent \hspace*{2em}"A run of sub-0.25 U deliveries totalling \{total\} U was not counted as a bolus.",\par
\noindent \hspace*{1em}],\par
\noindent \hspace*{1em}"small\_deliveries\_many": [\par
\noindent \hspace*{2em}"\{count\} runs of deliveries under 0.25 U per row (\{total\} U) were not counted as boluses.",\par
\noindent \hspace*{2em}"\{count\} runs of sub-0.25 U deliveries totalling \{total\} U were not counted as boluses.",\par
\noindent \hspace*{1em}],\par
\noindent \hspace*{1em}"basal\_zero\_ended": [\par
\noindent \hspace*{2em}"Basal delivery was zero for \{dur\} minutes, ending \{end\_m\} minutes before the decision.",\par
\noindent \hspace*{2em}"A \{dur\}-minute zero-basal stretch ended \{end\_m\} minutes before the decision.",\par
\noindent \hspace*{1em}],\par
\noindent \hspace*{1em}"basal\_zero\_ongoing": [\par
\noindent \hspace*{2em}"Basal delivery had been zero for \{dur\} minutes and still was in the slot before the decision.",\par
\noindent \hspace*{2em}"A \{dur\}-minute zero-basal stretch ran up to the slot before the decision.",\par
\noindent \hspace*{1em}],\par
\noindent \hspace*{1em}"basal\_changes\_one": [\par
\noindent \hspace*{2em}"The basal rate changed once, \{m\} minutes before the decision.",\par
\noindent \hspace*{2em}"One basal rate change, \{m\} minutes before the decision.",\par
\noindent \hspace*{1em}],\par
\noindent \hspace*{1em}"basal\_changes\_many": [\par
\noindent \hspace*{2em}"The basal rate changed \{n\} times, last \{m\} minutes before the decision.",\par
\noindent \hspace*{2em}"\{n\} basal rate changes, the latest \{m\} minutes before the decision.",\par
\noindent \hspace*{1em}],\par
\noindent \hspace*{1em}"temporal\_empty": [\par
\noindent \hspace*{2em}"No bolus, carbohydrate entry, small delivery run or basal change was recorded in \{period\}.",\par
\noindent \hspace*{2em}"\{period\_capitalized\} has no bolus, carbohydrate entry, small delivery run or basal change.",\par
\noindent \hspace*{1em}],\par
\noindent \}\par
\smallskip
\end{tcolorbox}
\begin{tcolorbox}[
  enhanced,
  breakable,
  sharp corners,
  colback=black!4,
  colframe=black,
  colbacktitle=black,
  coltitle=white,
  boxrule=0.4pt,
  title={GLOBAL},
  center title,
  fonttitle=\normalfont\small,
  fontupper=\fontsize{8.2}{10.2}\selectfont\ttfamily,
  left=7pt,right=7pt,top=5pt,bottom=7pt,
  boxsep=0pt,
  toptitle=5pt,bottomtitle=5pt,
  before skip=7pt,after skip=7pt,
  parbox=false
]
\raggedright\sloppy\setlength{\parindent}{0pt}\setlength{\parskip}{0pt}
\noindent CONTEXT\_CAPTION\_TEMPLATES = [\par
\noindent \hspace*{1em}"Presentation context includes \{supporting\_findings\}.",\par
\noindent \hspace*{1em}"Available contextual evidence includes \{supporting\_findings\}.",\par
\noindent \hspace*{1em}"The available clinical context mentions \{supporting\_findings\}.",\par
\noindent \hspace*{1em}"Before the decision point, the contextual record contains \{supporting\_findings\}.",\par
\noindent \hspace*{1em}"Contextual cues before the decision point include \{supporting\_findings\}.",\par
\noindent \hspace*{1em}"Relevant presentation and workflow context includes \{supporting\_findings\}.",\par
\noindent \hspace*{1em}"Earlier clinical context includes \{supporting\_findings\}.",\par
\noindent \hspace*{1em}"Pre-decision context includes \{supporting\_findings\}.",\par
\noindent ]\par
\smallskip
\noindent GLOBAL\_CAPTION\_TEMPLATES = [\par
\noindent \hspace*{1em}"Overall observed state before the decision: \{clinical\_state\}, supported by \{supporting\_findings\}.",\par
\noindent \hspace*{1em}"The pre-decision state is most consistent with \{clinical\_state\}; key evidence is \{supporting\_findings\}.",\par
\noindent \hspace*{1em}"Across directly observed context and signals, the dominant state is \{clinical\_state\}, with \{supporting\_findings\}.",\par
\noindent ]\par
\smallskip
\noindent CGM\_CONTEXT\_CAPTION\_TEMPLATES = \{\par
\noindent \hspace*{1em}"context": [\par
\noindent \hspace*{2em}"Patient context: \{items\}.",\par
\noindent \hspace*{2em}"Patient: \{items\}.",\par
\noindent \hspace*{1em}],\par
\noindent \hspace*{1em}"context\_empty": [\par
\noindent \hspace*{2em}"No patient context was recorded.",\par
\noindent \hspace*{2em}"Patient context is not recorded.",\par
\noindent \hspace*{1em}],\par
\noindent \}\par
\smallskip
\noindent CGM\_GLOBAL\_CAPTION\_TEMPLATES = \{\par
\noindent \hspace*{1em}"global\_states": [\par
\noindent \hspace*{2em}"Overall state: \{states\}.",\par
\noindent \hspace*{2em}"Overall pre-decision state: \{states\}.",\par
\noindent \hspace*{1em}],\par
\noindent \hspace*{1em}"global\_none": [\par
\noindent \hspace*{2em}"Overall state: no selected glycaemic state; direct pre-decision evidence was limited.",\par
\noindent \hspace*{2em}"Overall state: no glycaemic state was selected; direct pre-decision evidence was limited.",\par
\noindent \hspace*{1em}],\par
\noindent \}\par
\smallskip
\end{tcolorbox}
\begin{tcolorbox}[
  enhanced,
  breakable,
  sharp corners,
  colback=black!4,
  colframe=black,
  colbacktitle=black,
  coltitle=white,
  boxrule=0.4pt,
  title={ACTION},
  center title,
  fonttitle=\normalfont\small,
  fontupper=\fontsize{8.2}{10.2}\selectfont\ttfamily,
  left=7pt,right=7pt,top=5pt,bottom=7pt,
  boxsep=0pt,
  toptitle=5pt,bottomtitle=5pt,
  before skip=7pt,after skip=7pt,
  parbox=false
]
\raggedright\sloppy\setlength{\parindent}{0pt}\setlength{\parskip}{0pt}
\noindent TREATMENT\_ACTION\_EVIDENCE\_TEMPLATES = \{\par
\noindent \hspace*{1em}"supported": [\par
\noindent \hspace*{2em}"Observed findings relevant to \{action\_evidence\_domain\_phrase\} include \{action\_evidence\_clause\}.",\par
\noindent \hspace*{2em}"The available pre-decision evidence for \{action\_evidence\_domain\_phrase\} is \{action\_evidence\_clause\}.",\par
\noindent \hspace*{1em}],\par
\noindent \hspace*{1em}"secondary\_only\_evidence": [\par
\noindent \hspace*{2em}"The observed window includes \{action\_evidence\_clause\}; this is nonspecific supporting evidence for \{action\_evidence\_domain\_phrase\} rather than a complete indication.",\par
\noindent \hspace*{2em}"\{action\_evidence\_clause\} is relevant to \{action\_evidence\_domain\_phrase\}, although the available evidence is indirect or incomplete.",\par
\noindent \hspace*{1em}],\par
\noindent \hspace*{1em}"fallback\_visible\_evidence": [\par
\noindent \hspace*{2em}"Visible pre-decision findings include \{action\_evidence\_clause\}; together they provide indirect or mixed context for \{action\_evidence\_domain\_phrase\}, not a complete indication.",\par
\noindent \hspace*{2em}"The observed state contains \{action\_evidence\_clause\}; these facts are relevant background for \{action\_evidence\_domain\_phrase\}, although none is a direct domain-specific match.",\par
\noindent \hspace*{1em}],\par
\noindent \hspace*{1em}"label\_review\_context\_only": [\par
\noindent \hspace*{2em}"Visible pre-decision findings include \{action\_evidence\_clause\}; they are retained as clinical context, not as a direct indication for \{action\_evidence\_domain\_phrase\}.",\par
\noindent \hspace*{2em}"The available state shows \{action\_evidence\_clause\}; because this recorded treatment can represent context-specific care, these facts are described as background for \{action\_evidence\_domain\_phrase\} rather than a definitive rationale.",\par
\noindent \hspace*{1em}],\par
\noindent \hspace*{1em}"label\_review\_no\_visible\_context": [\par
\noindent \hspace*{2em}"The recorded treatment belongs to a context-dependent category; no reliable pre-decision fact is stated as a direct indication.",\par
\noindent \hspace*{2em}"This action category can reflect context-specific care, and the visible window does not support a label-specific rationale statement.",\par
\noindent \hspace*{1em}],\par
\noindent \hspace*{1em}"limited\_observable\_evidence": [\par
\noindent \hspace*{2em}"The available pre-decision record contains limited observable evidence specific to this action domain.",\par
\noindent \hspace*{2em}"No compatible state fact was selected for an action-specific evidence statement in this window.",\par
\noindent \hspace*{1em}],\par
\noindent \hspace*{1em}"no\_immediate\_treatment\_evidence": [\par
\noindent \hspace*{2em}"The observed pre-decision state does not assert a specific immediate treatment-level action from the available evidence.",\par
\noindent \hspace*{2em}"No immediate treatment domain is selected from the visible pre-decision context and signals.",\par
\noindent \hspace*{1em}],\par
\noindent \}\par
\smallskip
\noindent DIAGNOSTIC\_ACTION\_EVIDENCE\_TEMPLATES = \{\par
\noindent \hspace*{1em}"direct\_note\_indication": [\par
\noindent \hspace*{2em}"The pre-decision note documents this diagnostic indication: \{action\_evidence\_clause\}.",\par
\noindent \hspace*{2em}"A documented indication before T0 was: \{action\_evidence\_clause\}.",\par
\noindent \hspace*{1em}],\par
\noindent \hspace*{1em}"inferred\_indication": [\par
\noindent \hspace*{2em}"The \{action\_evidence\_domain\_phrase\} workup was likely prompted by \{action\_evidence\_clause\}; this is an inferred indication from pre-T0 evidence, not an explicitly documented reason.",\par
\noindent \hspace*{2em}"Pre-T0 findings suggest a possible indication for \{action\_evidence\_domain\_phrase\}: \{action\_evidence\_clause\}; the note did not state this rationale directly.",\par
\noindent \hspace*{1em}],\par
\noindent \hspace*{1em}"inferred\_general\_indication": [\par
\noindent \hspace*{2em}"The \{action\_evidence\_domain\_phrase\} workup was likely part of broader diagnostic assessment, but no specific pre-T0 indication was available.",\par
\noindent \hspace*{2em}"A diagnostic indication for \{action\_evidence\_domain\_phrase\} is plausible from the clinical setting, although the available record does not identify a specific reason.",\par
\noindent \hspace*{1em}],\par
\noindent \hspace*{1em}"supported": [\par
\noindent \hspace*{2em}"Diagnostic-workup evidence for \{action\_evidence\_domain\_phrase\} includes \{action\_evidence\_clause\}.",\par
\noindent \hspace*{2em}"The pre-decision record supports \{action\_evidence\_domain\_phrase\} through \{action\_evidence\_clause\}.",\par
\noindent \hspace*{1em}],\par
\noindent \hspace*{1em}"secondary\_only\_evidence": [\par
\noindent \hspace*{2em}"The observed window includes \{action\_evidence\_clause\}; this is indirect supporting context for \{action\_evidence\_domain\_phrase\}.",\par
\noindent \hspace*{2em}"Available context such as \{action\_evidence\_clause\} is relevant to \{action\_evidence\_domain\_phrase\}, but is not a complete indication.",\par
\noindent \hspace*{1em}],\par
\noindent \hspace*{1em}"fallback\_visible\_evidence": [\par
\noindent \hspace*{2em}"Visible pre-decision findings include \{action\_evidence\_clause\}; they provide broad context for \{action\_evidence\_domain\_phrase\}, not a definitive indication.",\par
\noindent \hspace*{2em}"The available state shows \{action\_evidence\_clause\}; its relationship to \{action\_evidence\_domain\_phrase\} remains indirect or mixed.",\par
\noindent \hspace*{1em}],\par
\noindent \hspace*{1em}"label\_review\_context\_only": [\par
\noindent \hspace*{2em}"Visible pre-decision findings include \{action\_evidence\_clause\}; they are clinical context for \{action\_evidence\_domain\_phrase\}, not a label-specific reason.",\par
\noindent \hspace*{2em}"The available state shows \{action\_evidence\_clause\}; the diagnostic action is retained without inferring an exact indication.",\par
\noindent \hspace*{1em}],\par
\noindent \hspace*{1em}"limited\_observable\_evidence": [\par
\noindent \hspace*{2em}"A diagnostic action was recorded, but no specific pre-decision indication was identifiable from the available context and signals.",\par
\noindent \hspace*{2em}"The visible window provides limited observable evidence for this diagnostic workup.",\par
\noindent \hspace*{1em}],\par
\noindent \hspace*{1em}"no\_immediate\_treatment\_evidence": [\par
\noindent \hspace*{2em}"No immediate treatment action is asserted; the observed decision context is diagnostic or otherwise non-treatment focused.",\par
\noindent \hspace*{2em}"The visible state does not establish an immediate treatment response, while diagnostic context may still be present.",\par
\noindent \hspace*{1em}],\par
\noindent \}\par
\smallskip
\noindent OR\_ACTION\_EVIDENCE\_TEMPLATES = \{\par
\noindent \hspace*{1em}"supported": [\par
\noindent \hspace*{2em}"Observed findings relevant to \{action\_evidence\_domain\_phrase\} include \{action\_evidence\_clause\}.",\par
\noindent \hspace*{2em}"The available pre-decision evidence for \{action\_evidence\_domain\_phrase\} is \{action\_evidence\_clause\}.",\par
\noindent \hspace*{2em}"For the current decision, visible evidence linked to \{action\_evidence\_domain\_phrase\} includes \{action\_evidence\_clause\}.",\par
\noindent \hspace*{2em}"Evidence aligned with \{action\_evidence\_domain\_phrase\} includes \{action\_evidence\_clause\}.",\par
\noindent \hspace*{2em}"The available state includes \{action\_evidence\_clause\}; these observations are relevant to \{action\_evidence\_domain\_phrase\}.",\par
\noindent \hspace*{2em}"Findings that bear on \{action\_evidence\_domain\_phrase\} are \{action\_evidence\_clause\}.",\par
\noindent \hspace*{1em}],\par
\noindent \hspace*{1em}"no\_immediate\_treatment\_evidence": [\par
\noindent \hspace*{2em}"No immediate action domain is selected from the visible pre-decision context and signals.",\par
\noindent \hspace*{2em}"The caption records the current state without assigning an immediate action domain.",\par
\noindent \hspace*{2em}"Current findings are described without linking them to a recorded immediate action.",\par
\noindent \hspace*{2em}"No action-specific rationale is asserted from the observed pre-decision state.",\par
\noindent \hspace*{2em}"The available pre-T0 state does not select a specific immediate response.",\par
\noindent \hspace*{1em}],\par
\noindent \}\par
\smallskip
\noindent CGM\_ACTION\_EVIDENCE\_TEMPLATES = \{\par
\noindent \hspace*{1em}"evidence\_for": [\par
\noindent \hspace*{2em}"Findings for an insulin bolus: \{findings\}.",\par
\noindent \hspace*{2em}"Evidence for a bolus: \{findings\}.",\par
\noindent \hspace*{1em}],\par
\noindent \hspace*{1em}"evidence\_against": [\par
\noindent \hspace*{2em}"Findings against an insulin bolus: \{findings\}.",\par
\noindent \hspace*{2em}"Evidence against a bolus: \{findings\}.",\par
\noindent \hspace*{1em}],\par
\noindent \hspace*{1em}"evidence\_none": [\par
\noindent \hspace*{2em}"No finding for or against an insulin bolus was selected from the observed window.",\par
\noindent \hspace*{2em}"No finding for or against a bolus was selected.",\par
\noindent \hspace*{1em}],\par
\noindent \}\par
\smallskip
\end{tcolorbox}

\endgroup

\end{document}